%% file: m9195.tex
\documentclass{ecai} 
\usepackage{latexsym}
\usepackage{amssymb}
\usepackage{amsmath}
\usepackage{amsthm}
\usepackage{booktabs}
\usepackage{enumitem}
\usepackage{graphicx}
\usepackage{color}

\usepackage{multirow}
\usepackage{threeparttable}

\usepackage{xcolor}
\usepackage{tikz}
\usetikzlibrary{calc, shapes.geometric, shapes.callouts, shapes.arrows, shapes.misc, positioning, fit, arrows.meta}
\usepackage[thicklines]{cancel}
\usepackage{listings}
\usepackage{pgfplots}
\pgfplotsset{compat=1.18}

\usepackage[nolist,nohyperlinks]{acronym}
\begin{acronym}
  \acro{LLM}{Large Language Model} 
  \acro{NLP}{Natural Language Processing}
  \acro{CoT}{Chain-of-thought}
  \acro{QA}{Question Answering}
  \acro{NL}{Natural Language}
  \acro{FOL}{First-Order Logic}
  \acro{AST}{Abstract Syntax Tree}
\end{acronym}

\usepackage{url}
\usepackage[utf8]{inputenc}
\usepackage{amsmath}
\usepackage{amsthm}
\usepackage{algorithm}
\usepackage{algorithmic}

\usepackage{fix-cm}

\newlength{\modelspacing}
\newcommand\Ccancel[2]{\renewcommand\CancelColor{\color{#1}}\cancel{#2}}

\begin{document}

%%%%%%%%%%%%%%%%%%%%%%%%%%%%%%%%%%%%%%%%%%%%%%%%%%%%%%%%%%%%%%%%%%%%%%%%

\begin{frontmatter}

%%% Use this command to specify your submission number.
%%% In doubleblind mode, it will be printed on the first page.

\paperid{9195} 

%%% Use this command to specify the title of your paper.

\title{Investigating the Robustness of Deductive Reasoning with Large Language Models}

%%% Use this combinations of commands to specify all authors of your 
%%% paper. Use \fnms{} and \snm{} to indicate everyone's first names 
%%% and surname. This will help the publisher with indexing the 
%%% proceedings. Please use a reasonable approximation in case your 
%%% name does not neatly split into "first names" and "surname".
%%% Specifying your ORCID digital identifier is optional. 
%%% Use the \thanks{} command to indicate one or more corresponding 
%%% authors and their email address(es). If so desired, you can specify
%%% author contributions using the \footnote{} command.

\author[A]{\fnms{Fabian}~\snm{Hoppe}\orcid{0000-0002-7047-2770}\thanks{Corresponding Author. Email: f.hoppe@vu.nl}}
\author[A]{\fnms{Filip}~\snm{Ilievski}\orcid{0000-0002-1735-0686}}
\author[B]{\fnms{Jan-Christoph}~\snm{Kalo}\orcid{0000-0002-5492-2292}} 

\address[A]{Vrije Universiteit Amsterdam}
\address[B]{Universiteit van Amsterdam}

%%% Use this environment to include an abstract of your paper.
\begin{abstract}
\input{sections/abstract} 
\end{abstract}

\end{frontmatter}

%%%%%%%%%%%%%%%%%%%%%%%%%%%%%%%%%%%%%%%%%%%%%%%%%%%%%%%%%%%%%%%%%%%%%%%%

\input{sections/introduction}
\input{sections/related_work}
\input{sections/noise}
\input{sections/method}
\input{sections/setup}
\input{sections/evaluation}
\input{sections/conclusion}

%%%%%%%%%%%%%%%%%%%%%%%%%%%%%%%%%%%%%%%%%%%%%%%%%%%%%%%%%%%%%%%%%%%%%%%%

%%% Use this environment to include acknowledgements (optional).
%%% This will be omitted in doubleblind mode.

\begin{ack}
This work has received funding from the European Union’s Horizon Europe research and innovation programme under the Marie Skłodowska-Curie grant agreement No 101073307. 
\end{ack}

%%%%%%%%%%%%%%%%%%%%%%%%%%%%%%%%%%%%%%%%%%%%%%%%%%%%%%%%%%%%%%%%%%%%%%%%

%%% Use this command to include your bibliography file.

\bibliography{bibliography}

% \clearpage
% \newpage
% \input{resubmission/resubmission}

\clearpage
\newpage
\appendix
\input{sections/appendix}

\end{document}

%% file: sections/abstract.tex
%% abstract.tex
%%
\acp{LLM} have been shown to achieve impressive results for many reasoning-based \ac{NLP} tasks, suggesting a degree of deductive reasoning capability. However, it remains unclear to which extent LLMs, in both informal and autoformalisation methods, are robust on logical deduction tasks.
Moreover, while many LLM-based deduction methods have been proposed, a systematic study that analyses the impact of their design components is lacking.
Addressing these two challenges, we propose \textit{the first study of the robustness of formal and informal LLM-based deductive reasoning methods}. We devise a framework with two families of perturbations: adversarial noise and counterfactual statements, which jointly generate seven perturbed datasets. We organize the landscape of LLM reasoners according to their reasoning format, formalisation syntax, and feedback for error recovery. 
The results show that adversarial noise affects autoformalisation, while counterfactual statements influence all approaches. 
Detailed feedback does not improve overall accuracy despite reducing syntax errors, pointing to the challenge of LLM-based methods to self-correct effectively. 

%% file: sections/introduction.tex
\section{Introduction}
\label{sec:introduction}

Deriving new knowledge from existing knowledge, as in deductive reasoning, is a key human cognitive skill necessary for various applications, including complex question-answering and decision-making. Deductive reasoning can be resource-intensive for humans (e.g., taking a lot of time), require specific expertise (e.g., logicians), and lead to incorrect conclusions (e.g., due to biases). This promotes automated deductive reasoning over \ac{NL} as a key objective of human-centric AI. Automatic deduction engines aim to support humans by providing certifiable reasoning chains, avoiding invalid inferences, and accelerating the process. To provide effective support for deductive reasoning, AI must be able to formalise knowledge and rules provided in \ac{NL} robustly.

Performing logical reasoning has received much interest in AI. In the early days, symbolic methods were aimed at transforming specific parts of language into logical statements~\cite{pereira1982logic}. Recently, \acp{LLM} have been shown to achieve impressive results for many reasoning-based \ac{NLP} tasks, suggesting a degree of deductive reasoning capability~\cite{srivastava2023beyond}. In particular, generating informal reasoning chains via \ac{CoT} prompting achieves good performance on many benchmarks~\cite{wei_chain_2022}. Contrary to symbolic methods, \acp{LLM} can answer a deductive reasoning task without providing a formal intermediate reasoning chain. Nevertheless, these informal reasoning chains do not need to follow truth-preserving proof rules, thus leading to chains that are hard to verify. Recent work shows that many informal reasoning chains suffer from a lack of \textit{faithfulness}~\cite{ye_unreliability_2022,tanneru_difficulty_2024}. 

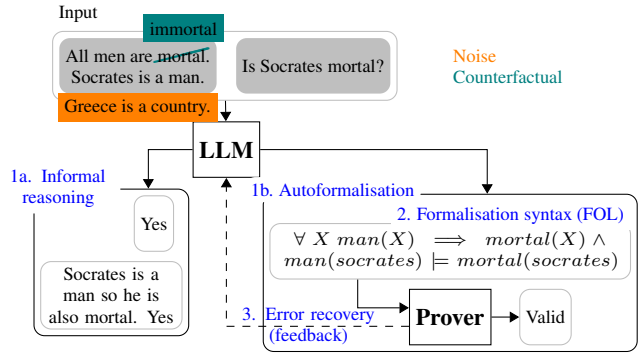
\begin{figure}[t]
    \centering
    \scriptsize
    \begin{tikzpicture}[node distance=.5em and .5em,
        every node/.style={align=center, minimum height=3em, minimum width=2em},
        label_style/.style={rectangle, minimum height=1em, rounded corners, fill=white}
        ]
        \node[draw](llm){\small \bfseries{LLM}};
        
        \node[rectangle, rounded corners, fill= lightgray, above= 4em of llm, anchor=east, xshift=-0.5em](input_context){All men are \Ccancel{teal}{mortal}. \\ Socrates is a man.};
        \node[rectangle, rounded corners, fill=lightgray, right= 1em of input_context](input_question){Is Socrates mortal?};
       
        \node[rectangle, rounded corners,draw=lightgray, fit={(input_context)(input_question)}](input){};
        \node[label_style, above=2.8em of input.west, anchor=west](input_label){ Input};
        
        \node[draw=lightgray, rectangle, rounded corners, below= 4.5em of llm.west, anchor=east, xshift=-1em](direct_answer){Yes};
        \node[draw=lightgray, rectangle, rounded corners, text width= 0.2\linewidth, below= of direct_answer, xshift=-2.3em](cot_answer){Socrates is a man so he is also mortal. Yes};
        \node[rectangle, rounded corners,draw, fit={(direct_answer)(cot_answer)}](informal){};
        \node[label_style, text width= 6em, above=4em of informal.west, anchor=west, xshift=-2em](informal_label){\textcolor{blue}{1a. Informal reasoning}};
        
        \node[draw=lightgray, rectangle, rounded corners, text width= 0.54\linewidth, below= 6 em of llm.east, anchor=west, xshift=0.5em](lf_text){$\forall~X~man(X) \implies mortal(X) \wedge man(socrates) \models mortal(socrates)$};
        \node[fill=white, rectangle, minimum height=1.5em, above=-0.4em of lf_text.north east, anchor=south east](fol){\textcolor{blue}{2. Formalisation syntax (FOL)}};
        \node[draw, below= of lf_text](atp){\small \bfseries{Prover}};
        \node[draw=lightgray, rectangle, rounded corners, right= 1.5em of atp](af_answer){Valid};
        \node[rectangle, rounded corners,draw, fit={(fol)(lf_text)(atp)(af_answer)}](auto){};
        \node[label_style, above=4.5em of auto.west, anchor=west, xshift=-1em](auto_label){\textcolor{blue}{1b. Autoformalisation}};

        \draw[-Triangle] (input)   -- (llm);

        \draw[-Triangle] (llm.west)   -| ([xshift=2em]informal.north);
        \draw[-Triangle] (llm.east)   -| ([xshift=2em]auto.north);

        \draw[-Triangle] ([xshift=-5em]lf_text.south)   |- ([yshift=.5em]atp.west);
        \draw[-Triangle] (atp.east)   -- (af_answer.west);

        \draw[dashed,-Triangle] ([yshift=-.5em]atp.west)   -| (llm.south) node [near start, xshift=-0.5em, text width=12em] {\textcolor{blue}{3. Error recovery \\ (feedback)}};

        \node[align=left, right=3em  of input_question.east, anchor=west](legend){ \textcolor{orange}{Noise}\\ \textcolor{teal}{Counterfactual}};
        \node[rectangle, fill=orange,  opacity=.75, overlay, minimum height=1.5em, below=-0.1em  of input_context.south, anchor=north](dis_example) {Greece is a country.};
        \node[rectangle, fill=teal, opacity=.75, overlay, minimum height=1.5em, above= of input_context.north east, xshift=-4em, anchor=west]  {immortal};
        
    \end{tikzpicture}
    \caption{Overview of our methodology for investigating the robustness of reasoning with \acp{LLM}. Our perturbations (noise and counterfactuals) are shown in orange and teal, respectively. The three dimensions of our \ac{LLM}-based methodological framework (reasoning format, syntax, and error recovery mechanism) are shown in blue.}
    \label{fig:overview}
\end{figure}

Addressing these challenges, \emph{autoformalisation} approaches ~\cite{liangming_pan_logiclm_2023,olausson_linc_2023} use \acp{LLM} to translate \ac{NL} input into a logical form, and a rigorous symbolic solver to perform the deductive reasoning. Autoformalisation is thus a hybrid approach, which aims to provide a faithful and verifiable reasoning chain while leveraging the linguistic manipulation skills of \acp{LLM}. 
Autoformalisation faces two key challenges: First, since they translate rich \ac{NL} into a limited grammar of symbols and operations, it is critical to leverage a syntax with an optimal tradeoff between translation accuracy and expressivity. Second, while autoformalisation chains provide an opportunity for syntactic and semantic validation and error analysis, designing an effective and efficient error recovery mechanism is non-trivial. While prior autoformalisation systems leverage multiple syntaxes and error recovery mechanisms, \textit{no systematic study has investigated their impact on autoformalisation accuracy}.

Meanwhile, a key requirement of \ac{LLM}-based reasoning methods is their robustness to noise~\cite{ebrahimi-etal-2018-hotflip} and out-of-distribution~\cite{hendrycks2021measuring} inputs. Given the strong performance of \acp{LLM} across many domains and benchmarks~\cite{sarlin2020superglue}, dealing with noisy data in reasoning has been considered more important~\cite{sourati-etal-2024-robust}. Most evaluations have focused on adversarial noise (e.g., lexical perturbations)~\cite{sarlin2020superglue}, while a recent study has also experimented with counterfactual statements~\cite{liu2023recall}. While robustness evaluations for \ac{NLP} tasks have yielded mixed results~\cite{wang_rupbench_2024,liu2023recall}, it remains \textit{unclear to which extent \acp{LLM}, in both informal and autoformalisation methods, are robust in logical deductions}.

We address these two challenges by \textbf{investigating the robustness of \ac{LLM}-based deductive reasoning methods} with the goal to advance the understanding of \ac{LLM}-based reasoning and shed light on methodological aspects to inform more complex, future reasoning systems. Our overall approach is summarised in Figure \ref{fig:overview}. 

Our study makes three contributions.
First, following standard practices in evaluating robustness, we devise a \textbf{robustness framework for logical deduction} with two families of perturbations: adversarial noise, where the model needs to preserve its label in the face of added irrelevant information, and counterfactual perturbations, where a single alteration in the context flips the label of the question. The combinations of the perturbations produce seven variants from a given dataset. The perturbed data is made available~\cite {hoppe_investigating_2025}.
Second, we synthesise the landscape of existing \ac{LLM}-based logical deduction methods into a \textbf{methodological framework with three dimensions}: reasoning format, grammar syntax, and error recovery mechanism. For each of these dimensions, we incorporate representative approaches in the literature and release the source code~\cite{hoppe_investigating_2025}.
Third, we perform \textbf{extensive experiments} with seven \acp{LLM} on eight perturbed variants of a recent modular benchmark. Our findings provide nuanced insights into the robustness of \ac{LLM}-based reasoning methods. 

%% file: sections/related_work.tex
\section{Related Work}

This section gives an overview of \ac{LLM}-based reasoning methods and studies that evaluate their robustness. 

\subsection{Methods for LLM-based Reasoning}
\paragraph{Informal reasoning.} 
Scaling up the size of \acp{LLM} enables strong performance in many \ac{NLP} tasks by few-shot prompting~\cite{brown_language_2020}, which suggests inherent reasoning capabilities.  
\acf{CoT} combines few-shot prompting with generating intermediate informal reasoning chains~\cite{wei_chain_2022}. 
These informal reasoning skills motivated more elaborate prompting techniques, like Zero-Shot \ac{CoT}~\cite{kojima_large_2022} or self-consistency by generating multiple chains~\cite{wang_selfconsistency_2022}, as well as more complex structures than chains, such as Tree of Thoughts~\cite{yao_tree_2023} and Graph of Thoughts~\cite{besta_graph_2024}. 
These methods use an \ac{LLM} to generate intermediate steps and evaluate the output through self-refinement.
Similarly, ProofWriter~\cite{tafjord_proofwriter_2021} improves multi-hop reasoning by adding the intermediate results to the reasoning context~\cite{tafjord_proofwriter_2021}.  
A key benefit of informal reasoning chains is their flexibility, but this comes at the expense of guaranteeing faithfulness. Consequently, methods combining \ac{LLM}s with formal reasoning have been suggested.

\noindent
\paragraph{Autoformalisation.} 
Instead of informal reasoning, another way combines an \ac{LLM} with a deterministic symbolic solver, e.g., a theorem prover. 
Here, the prover guarantees faithful and deterministic reasoning.
This combination is known as \emph{autoformalisation}.
One of the first autoformalisation approaches using formal reasoning chains in combination with prompting is PAL~\cite{gao_pal_2023}. 
The authors generate Python snippets alongside informal steps and generate the final response by executing the generated code snippets.  
Logic-LM~\cite{liangming_pan_logiclm_2023} prompts \acp{LLM} to generate multiple task-specific formalisations (logic programming, first-order logic, satisfiability modulo theories and constraint satisfaction), which are solved by dedicated solvers. 
They report higher robustness for longer reasoning chains compared to \ac{CoT} reasoning.
The extension Logic-LM++~\cite{kirtania_logic-lm_2024} tries to avoid new syntax errors by integrating a self-refinement mechanism.
The LINC~\cite{olausson_linc_2023} approach uses the idea of self-consistency from \ac{CoT} and generates multiple formalisations to avoid formalisation errors. 
Autoformalisation models achieve high accuracy for many deductive reasoning benchmarks, showing clear benefits for complex reasoning.

\paragraph{Comparison.} Our work is the first to explore the robustness of \ac{LLM} reasoning approaches from prior work: direct few-shot prompting, \ac{CoT}, and autoformalisation. Another contribution of our work is consolidating these methods with their syntax and error recovery choices into a coherent methodological framework.

\subsection{Robustness Evaluation of LLM Reasoning}
\paragraph{Evaluating deductive reasoning.} The improvements of \ac{LLM}-based reasoning have inspired the development of benchmarks investigating capabilities for deductive reasoning. 
The synthetic PrOntoQA~\cite{saparov_language_2022} dataset is built on modus ponens multi-hop reasoning and confirms reasoning capabilities for the largest models. 
However, they noted issues with proof planning and selecting the correct proof steps for longer reasoning chains. 
The FOLIO~\cite{han_folio_2022} and AR-LSAT~\cite{zhong_analytical_2022} benchmarks confirmed these errors for more complex reasoning and more naturalistic language. 
LogicBench~\cite{mihir_parmar_logicbench_2023} is a recent benchmark that systematically studies the performance of \ac{LLM}-based reasoners across multiple inference rules (e.g., modus ponens), reporting a good model performance for predicate logic and first-order logic.
%% Unfaithful reasoning  and spurious correlation 
A vital challenge identified by these works is unfaithful reasoning, i.e., the intermediate steps are not used to infer the final result~\cite{ye_unreliability_2022,lanham_measuring_2023,tanneru_difficulty_2024}.

\paragraph{Robustness studies.}
A key requirement of \ac{LLM}-based reasoning methods is their robustness to noise~\cite{ebrahimi-etal-2018-hotflip} and out-of-distribution~\cite{hendrycks2021measuring} inputs. Given the strong performance of \acp{LLM} across many domains and benchmarks~\cite{sarlin2020superglue}, dealing with noisy data in reasoning tasks has been considered more important~\cite{sourati-etal-2024-robust}, including adversarial and counterfactual perturbations.
A variety of robustness tests have therefore been designed~\cite{wang_measure_2022}. 
These robustness benchmarks rely on the original problem's perturbations through paraphrasing and distractions on character, word, and sentence levels~\cite{sourati-etal-2024-robust,sarlin2020superglue}. 
Many works try to generate adversarial examples to better understand the generalisation capabilities of \acp{LLM}, which are shown to significantly decrease \ac{LLM} performance~\cite{wang_measure_2022}.
RUPbench is a recent robustness study on logical reasoning of \acp{LLM}~\cite{wang_rupbench_2024}, covering many reasoning datasets and all three types of perturbations.
Another category involves semantic changes in the input texts. 
Semantic changes can also be performed purely on a linguistic or logical level.
Recent approaches use these logic-based perturbations~\cite{nakamura-etal-2023-logicattack}, which align with our robustness framework. Meanwhile, counterfactual studies of \ac{LLM} robustness have been less common. One exception is RECALL~\cite{liu2023recall}, a benchmark based on external knowledge bases, whose study reveals that \acp{LLM} are generally susceptible to external knowledge with counterfactual information and that simple mitigation strategies cannot significantly alleviate this challenge.

\paragraph{Comparison.} We conduct the first investigation of robustness differences between formal and informal \ac{LLM}-based logical deduction methods. For this purpose, we base our experiments on LogicBench~\cite{mihir_parmar_logicbench_2023}, which systematically includes nine inference rules mapped to natural language situations. We develop seven additional variants of LogicBench incorporating adversarial noise and counterfactual statements, in line with prior work like \cite{nakamura-etal-2023-logicattack} that studies \ac{LLM} robustness on other reasoning tasks.

%% file: sections/noise.tex
\section{Framework for Evaluating Robustness in Logical Deduction Tasks}
\label{sec:robust}
\textbf{Task definition.} Deductive reasoning is commonly formatted as binary \ac{QA} over a given context. 

The task input consists of a \ac{NL} question $q$ and a set of logical premises transformed into an \ac{NL} context $c$. The output of the task $a$ is one of the two possible answers: true or false, indicating whether the context supports or refutes the question. % $\{true,false\}$.

Formally, an \ac{NL} deductive reasoning task defines a function $f: (c,q) \rightarrow \{true, false\}$. 

\noindent \textbf{Robustness types.} Following prior work on investigating the robustness of \ac{LLM}s~\cite{sarlin2020superglue,liu2023recall}, we conceptualize two families of context perturbations.
First, inspired by adversarial robustness studies~\cite{nakamura-etal-2023-logicattack}, we posit that a reasoning system must be robust to adversarial \textbf{noise} (distractions), i.e., including irrelevant information in the context should not alter its prediction. 
Noise can shift the focus of a text, making it more difficult for an \ac{LLM} to capture the relevant context.
Second, a reasoning system must be robust to \textbf{counterfactual} shifts in the context. Namely, the system must be faithful to the provided context without including biases from implicit world models~\cite{liu2023recall}. 
If the context states that \emph{men are immortal}, the reasoning must overwrite its belief of men as mortal. 
By introducing counterfactual perturbations contradicting common sense, we investigate whether \ac{LLM}-based logical deduction methods use reasoning shortcuts or perform genuine reasoning.
We focus our perturbations on the \ac{NL} context $c$ rather than the question $q$ because it corresponds to background knowledge sources, which tend to vary significantly in real-world applications (e.g., compare reasoning based on a research article to reasoning over a list of facts). The perturbed task can be formalised as $f': (c',q) \rightarrow \{true, false\}$. 

Next, we detail our framework for evaluating the robustness of \ac{LLM}-based deduction methods to noise and counterfactual shifts.

\subsection{Adversarial Noise}
\textbf{Formalisation.} As monotonic reasoning, deductive reasoning must remain invariant to newly added irrelevant information, i.e., additional text that does not change the semantics of the existing premises does not change the conclusion. Let us consider a perturbed context that includes noisy information: $c' = d_1 \dots d_k c$, where each $d_i$ denotes a noisy sentence concatenated to $c$ for $k \in \{1,2,4\}$. The task function $f'$ must resolve to the same output as the original $f$ for its proof to remain valid. To avoid any distractions that might change the original semantics (e.g., by breaking inter-sentence co-references), we append distractions only to the beginning of the context.  

\noindent \textbf{Design of adversarial noise.}
We define three types of noise relevant to logical deduction tasks, which vary in their degree of referential content, formalisation complexity, and depth of logical reasoning. All noise sentences are sampled in a way that guarantees they do not impact the semantics of the context and the original proof. Figure \ref{fig:perturbation_example} (top-right) shows examples of the three noise types.

\paragraph{1.} \textbf{Encyclopedic (E)} perturbations are \ac{NL} sentences expressing factual information such as \textit{It is 21 km from Karimnagar, on the highway from Karimnagar to Peddapa}. They express real-world information following pragmatic principles of language~\cite{grice1975logic}.
Encyclopedic sentences are often difficult or even impossible to formalise in first-order logic. At the same time, encyclopedic facts are not connected by complex logical relations and lack the linguistic structure typical for reasoning contexts. Consequently, this perturbations only has a low semantic similarity to the original context.  In summary, they represent world information, have a high formalisation complexity, and have low logical reasoning depth.
\paragraph{2.} \textbf{Logical (L)} statements provide a typical structure of reasoning contexts and contain only knowledge that can be natively formalised, such as \textit{All dresses are clothes}. Logical sentences include information about the world, albeit in a fictional form. As such, the semantic similarity is larger than encyclopedic perturbations due to the shared typical structure. 
The formalisation usually requires more complex reasoning, like multi-hop inferences. Thus, logical perturbations introduce limited world information, have low formalisation complexity, and high reasoning depth.
\paragraph{3.} \textbf{Tautological (T)} perturbations are easily recognisable correct statements, e.g., \emph{True is not false}. They may include negations and one disjunction or conjunction. These perturbations provide the highest semantic similarity to the original context because they only contain the shared typical structure of the reasoning context without including information about the world. As such, they contain no referential information and require simple reasoning. However, their formalisation is often difficult because many first-order logic syntaxes do not consider predefined truth constants, like $\bot$, $\top$. In summary, tautologies in \ac{NL} contain no referential information, have high formalisation complexity, and have low reasoning depth.

\begin{figure}[t]
    \centering
    \scriptsize
    \begin{tikzpicture}[node distance=1em and 0.5em, 
        every node/.style={align=left},
        label_style/.style={rectangle, rounded corners, fill=white}
        ]
        \node[text width=0.35\linewidth, rectangle, rounded corners, fill=lightgray] (original) {If an individual drinks water, they will be hydrated.};
        \node[label_style, above= 0.5em of original.north west, anchor=west, draw=gray]{ Original (O)};

        \node[draw, text width=0.275\linewidth, below= 2.5em of original.south west, anchor=north west, xshift=2.1em] (folio) {All dresses are clothes. If an individual ...};
        \node[label_style, above= 0.5em of folio.north west, anchor=west](folio_label){Logical (L)};
        \node[draw, text width=0.8\linewidth, below= 1.5em of folio.south west, anchor=north west] (tautology) {Not true or false is not true. If an individual ...};
        \node[label_style, above=0.5em of tautology.north west, anchor=west]{Tautological (T)};
        \node[draw, text width=0.8\linewidth, below= 1.5em of tautology.south west, anchor=north west] (wiki) {It is 21 km from Karimnagar, on the highway from Karimnagar to Peddapalli. If an individual ...};
        \node[label_style, above=0.55em of wiki.north west, anchor=west]{Encyclopedic (E)};
        \node[rotate=90, left= 1em of folio, xshift=1.5em] (label_distractions) {\textbf{Adversarial noise}};
        
        \node[draw, text width=0.475\linewidth, right= of original] (original_counter) {If an individual drinks water, they will not be hydrated.};
        \node[label_style, above= 0.5em of original_counter.north west, anchor=west]{Original ($\text{O}_C$)};
        \node[draw, label_style, below= 1.1em of original_counter, xshift=-3.5em](folio_counter){ Logical ($\text{L}_C$)};
         \node[draw, label_style, right=of folio_counter](tautology_counter){Tautological ($\text{T}_C$)};
        \node[draw, label_style, below= 0.1em of folio_counter, xshift=2.5em](wiki_counter){Encyclopedic ($\text{E}_C$)};
  
        \node[above= 0.7em of original_counter] (counter_label) {\textbf{Counterfactual}};
        \node[draw,  fit={(counter_label)(original_counter)(tautology_counter)(wiki_counter)} ] (counter) {}; 
        \node[draw,  fit={(label_distractions)(folio_label)(wiki)(tautology_counter)(tautology)} ] (distractions) {}; 
    \end{tikzpicture}
    \caption{Example of a premise and its perturbations.}
    \label{fig:perturbation_example}
\end{figure}

\subsection{Counterfactual Shifts}

\textbf{Formalisation.}
We consider common-sense contradictions by altering original statements in $c$ into counterfactual ones. The inclusion of counterfactual statements is motivated by the requirement for faithful reasoning. Namely, the validity of a deduction depends only on the structure of the logical form, which in turn should follow the original task description in the context and the question. 
Therefore, the reasoning capabilities of a model should not rely on prior knowledge. Instead, explicitly stated premises should be prioritised over implicit model knowledge and overwrite it. This behaviour allows models to reason over hypothetical and counterfactual premises, enabling the derivation of new (scientific) knowledge, e.g., deriving non-Euclidean geometry by replacing the parallel postulate, or science fiction authors to create complete, logically consistent alternative worlds based on one changed premise. Moreover, it ensures a more controlled evaluation setup for evaluating reasoning independently of the quality of the internal model knowledge, since all models use the same premises.
To introduce counterfactual premises, we do not add new sentences; instead, we negate sentences from the original context $c$. Since original premises often state common-sense knowledge, their negation naturally contradicts world knowledge. For example, in the modus ponens inference in Figure \ref{fig:overview}, we negate $mortal(X)$, stating that \emph{All men are immortal}. The altered context $c'$ is formalised as follows: $c'=neg(c)$, where $neg(c)$ negates one of the terms in the original context.
Naturally, these perturbations influence the inferred answer and depend on the applied inference rule. The negated term is selected so that the label of the resulting function $f'$ is opposite from that of the original function $f$.

\noindent \textbf{Design of counterfactual perturbations.} 
We assume a deductive reasoning dataset where the natural language form corresponds to a logical formula. Then, we introduce counterfactual perturbations by altering each logical formula using predefined rules that negate the formula (see the full list of rules in the Appendix).
For example, we negate the consequent $\mathsf{q}(a)$ for the first implication of a constructive dilemma and adapt the inference to $\neg \mathsf{q}(a) \vee \mathsf{s}(a)$. 
Then, the negation to create the context $c'$ is manually added before the relevant terms in $c$, thus guaranteeing high data quality. 
Since the inference results in the opposite label for $f'$, we adapt the target label accordingly. Figure \ref{fig:perturbation_example} (top-right) shows an example of such a contradiction. Importantly, all counterfactual perturbations can be combined with noise perturbations, leading to counterfactual versions of the original ($O_C$), encyclopedic ($E_C$), logical ($L_C$), and tautological ($L_T$) sets.

%% file: sections/method.tex
\section{Methodological Framework}
We consider three key design dimensions: reasoning format, syntax, and error recovery mechanism to systematically analyse robust deductive reasoning capabilities for \ac{LLM}-based methods. We describe each of these dimensions and their representative approaches studied in this work. 

\subsection{Reasoning Format}
\ac{LLM}-based methods can either: a) operate without any explicit formalisation as informal reasoning systems and generate the answer based on their internal representation, or b) generate a formal representation of the given input, which is fed into a theorem prover to find a valid proof.

The \textbf{informal} reasoning method instructs an \ac{LLM} to answer $q$ with \emph{Yes} or \emph{No} based on a context.
We employ few-shot prompting using three manually engineered in-context examples. To make the model more robust, two of the three in-context examples include distractions with irrelevant information. We explicitly note in the instruction part of the prompt that the provided contexts may contain irrelevant details.
The model is evaluated in two modes: \textit{direct} prompting, where it answers directly, and \textit{\ac{CoT}} prompting, where it generates step-by-step reasoning in natural language before generating an answer. To support \ac{CoT} prompting, we manually created in-context examples with fine-grained natural language reasoning steps designed to improve performance~\cite{wei_chain_2022}. 
The model’s answers are extracted using a regular expression (details in  Appendix~\ref{app:prompts}).

We include an \textbf{autoformalisation} method combining a symbolic theorem prover with an \ac{LLM}. This approach follows Logic-LM~\cite{liangming_pan_logiclm_2023} to formalise the context and query into symbolic representations, which are then automatically proven. 
The process is divided into two subtasks: First, the model generates formal representations of the context and query, referred to as $c_{LF}$ (context logical form) and $q_{LF}$ (query logical form). Second, a symbolic theorem prover evaluates these logical forms to determine whether $q_{LF}$ can be derived from $c_{LF}$, producing a \emph{true} or \emph{false} outcome. This approach allows for a transparent and verifiable reasoning process grounded in logical consistency.
In practice, we prompt the \ac{LLM} to create the logical forms using the same three in-context examples as in the informal reasoning approach. The prompt is extended with instructions describing the formalisation syntax, following the methodology of Logic-LM. The resulting logical forms are parsed and combined into an \ac{AST}, which provides the input for a theorem prover.

\subsection{Formalisation Syntax}
\label{subsec:syntax}

\begin{figure}[t]
    \centering
    \scriptsize
    \begin{tikzpicture}[node distance=0.2em and 0.2em,
        every node/.style={align=left,},
        fnode/.style={draw, },
        label_style/.style={rectangle, rounded corners, fill=white}
        ]        
        \node[text width=0.825\linewidth, rectangle, rounded corners, fill=lightgray,] (NL) {If an individual drinks water, they will be hydrated.};
        
        \node[fnode, text width=0.4\linewidth, below= 3.3em of NL.west, anchor=west] (FOL) {$\forall~x~drinkWater(x) \implies hydrated(x)$};
        \node[label_style, above= 1.8em of FOL.west, anchor=west]{FOL};
        
        \node[fnode, text width=0.4\linewidth, right= of FOL, anchor=west] (RFOL) {$\forall~?x~drinkWater(?x) \implies hydrated(?x)$};
        \node[label_style, above= 1.8em of RFOL.west, anchor=west]{R-FOL};

        \node[fnode, text width=0.825\linewidth, below= 3.6em of FOL.west, anchor=west] (TPTP) { \ttfamily fof(a0,axiom,![X]:drinkWater(X) => hydrated(X)). }; 
        \node[label_style, above= 1.5em of TPTP.west, anchor=west]{TPTP};
    \end{tikzpicture}
    \caption{Examples of the three syntaxes: FOL, R-FOL, and TPTP.}
    \label{fig:formalisation_example}
\end{figure}
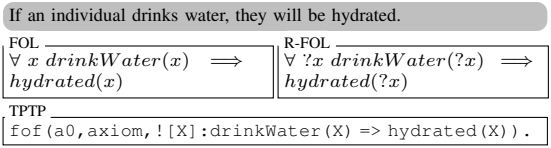

While the reasoning performance should be independent of the particular formalisation syntax, models may perform better with specific syntaxes, e.g., because of their frequency in the training data~\cite{razeghi2022impact}.
It is an open question whether the choice of syntax for formal reasoning impacts the model's translation performance and the robustness of its reasoning. Although all the syntaxes we consider represent \ac{FOL}, their surface form variations may influence the formalisation ability of \acp{LLM}.
The three evaluated syntaxes, illustrated in Figure~\ref{fig:formalisation_example}, contain identical information and are interchangeable, which ensures flexibility and allows the framework to include other syntaxes in the future:

\paragraph{FOL} is widely used in logic classes and academic papers. It incorporates mathematical symbols such as $\forall$ and $\exists$ and implicitly distinguishes between variables and individuals. This syntax is employed by Logic-LM~\cite{liangming_pan_logiclm_2023}.
\paragraph{R-FOL} is a variation of \ac{FOL} that explicitly differentiates variables and individuals by requiring variables to start with a question mark. This \emph{resolves} the syntax ambiguity in \ac{FOL}. 
\paragraph{TPTP} as the abbreviation of \emph{Thousands of Problems for Theorem Provers} is a Prolog-like formalisation language developed for theorem provers. It avoids mathematical symbols and mandates that variables begin with an uppercase letter. While TPTP supports higher-order logic, we limit our scope to its first-order fragment (fof), using a syntax derived from its official specification~\cite{Sut24}.

\subsection{Error Recovery Mechanism}
To make autoformalisation more robust, we synthesise strategies for handling syntactic and semantic errors. Syntactic errors occur when the logical forms generated by \acp{LLM} do not follow the required syntax. Syntactic errors are easy to detect as logical forms cannot be parsed if they violate grammatical rules. In contrast, semantic errors, such as incomplete context representation, are more challenging to identify and resolve. We apply task-specific heuristics to identify suspicious constructs that lead to semantic errors and generate warnings. Unknown predicates as part of $q_{LF}$ are one example of such a construct, because they indicate an incomplete context. 
We consider four strategies for handling these errors:

\paragraph{No recovery.} The baseline approach does not attempt to correct errors. Instead, we predict a random value, \emph{true} or \emph{false}, as a fallback strategy. We avoid introducing an evaluation bias associated with more complex strategies, such as \ac{CoT}-based refinement, which is commonly done in prior work~\cite{liangming_pan_logiclm_2023,kirtania_logic-lm_2024}, as this would blur the comparison with other methods.
\paragraph{Error type feedback.} The \ac{LLM} is prompted to refine the logical form using a generic parsing error message, such as \emph{'parsing error'}. This type of message does not need a parser with an error handler, though it fails to point to specific errors in the logical form. 
Prior work has shown the effectiveness of this feedback method~\cite{liangming_pan_logiclm_2023}.
\paragraph{Error message feedback.} A more detailed approach where the \ac{LLM} is given specific feedback, highlighting the exact parts of the logical form that violate the syntax. Creating this kind of feedback necessitates an error-handling strategy for the parser. For example, the missing argument in $man \wedge mortal(Socrates)$ results in the error message: \emph{mismatched input '$\wedge$' expecting '('}.
Using error messages from parsers as feedback to improve \acp{LLM} performance has shown promising results in code generation~\cite{zhang-etal-2023-self,jiang-etal-2024-leanreasoner}.
\paragraph{Warning feedback.} An extension of the error messages with warnings generated from heuristics to recognise semantic errors, inspired by the ``soft'' critics in the LLM-modulo method~\cite{kambhampati_llms_2024}, where it has been shown to enhance robustness. Notably, soft critics have not been incorporated in prior \ac{LLM}-based methods for logical deduction.

%% file: sections/setup.tex
\section{Experimental Setup}
The experimental setup, including the generated perturbed data, the source code and configuration files, is made available on Zenodo~\cite{hoppe_investigating_2025} and GitHub\footnote{\url{https://github.com/Fab-Hop/langdeductive}} to facilitate reproducibility. 
\paragraph{Dataset details.}
The robustness evaluation is based on the \ac{FOL} part of the LogicBench (Eval)~\cite{mihir_parmar_logicbench_2023} dataset, containing $520$ samples with $180$ unique contexts. LogicBench systematically covers nine different inference rules. The samples are automatically generated by prompting GPT-3.5 and manually verified. $340$ of the total of $520$ samples are negative examples constructed by negating the conclusion. If the conclusion is a disjunction, each part of the disjunction is negated, resulting in a slight class imbalance. The authors report a mean accuracy of around $90\%$ for three human annotators. By applying our robustness framework from §\ref{sec:robust}, we obtain seven perturbed variants of LogicBench.

\paragraph{Perturbations.}
The noise sentences are randomly sampled from a source $s$. We sample encyclopedic perturbation sentences from $10,000$ abstracts of Wikipedia articles gathered via its API.
As a logical reasoning source, we use sentences from $1001$ contexts of the deduction \ac{QA} benchmark FOLIO~\cite{han_folio_2022}. 
As tautologies, we manually write $22$ sentences. All sentences use negations and, at most, one disjunction or conjunction. A complete list can be found in  Appendix~\ref{app:tautology}.
We randomly sample noise perturbations and add them to each sample. We do not alter the class distribution, i.e., we keep LogicBench's original class imbalance.
The three types of noise sources have varying, yet consistently low, levels of semantic relevance, quantified as an average cosine similarity of sentence-transformers embeddings between the added noise and existing context. Tautological noise is most similar ($0.20$), followed by logical noise ($0.09$), and encyclopedic noise ($0.04$). The low semantic similarity and the careful construction of the dataset, i.e. appending distractions only at the beginning of the sentence ensure no interference with the original semantics. 

We consider unique contexts for eight out of the nine logical forms to create counterfactual statement perturbations, resulting in $160$ samples. We create a balanced dataset by alternating between valid and invalid queries from LogicBench. This dataset is a synthetic dataset created by an \ac{LLM} using its latent knowledge. Thus, the original premises align with world knowledge encoded in an \ac{LLM} and the negation of a premise creates a counterfactual premise adapted to this latent knowledge. 

\paragraph{Metrics.} We use \emph{accuracy} as a standard metric for classification tasks. We report \emph{execution rate} as the fraction of parsable texts and \emph{valid accuracy} as the accuracy on these parsable samples in the Appendix~\ref{app:experiments} due to space limitations. 

\paragraph{LLMs.} 
We test GPT 4o-mini, as well as a smaller and a larger variant for the three open-source \ac{LLM} families: Gemma-2 (9b and 27b), Mistral (7b and Small), and Llama 3.1 (8b and 70b). 
If available, the tested \acp{LLM} are used in their instruction-tuned variant. Except for GPT 4o-mini, all models are provided from hugging-face and used with 4-bit quantisation. 
Our experiments use a default temperature of $1.0$ to keep the output distributions unmodified. Furthermore, we verify the consistency of our findings by conducting experiments with a temperature setting of $0$.

\paragraph{Syntax Parsers.}
For each grammar in §~\ref{subsec:syntax}, we generate a parser using ANTLR4\footnote{\url{https://www.antlr.org/}} to decouple the logical forms from the input requirements of the theorem prover. The FOL variant is adapted from the ANTLR grammars-v4 repository\footnote{\url{https://github.com/antlr/grammars-v4}}. The parser validates the syntax and, if necessary, translates it into TPTP to pass on to the theorem prover. The Vampire theorem prover\footnote{\url{https://vprover.github.io}} processes this formal method's input, generating proofs by refutation to check whether the query logically follows from the context.

The syntax parser generates specific feedback for all error recovery strategies to prompt the \ac{LLM}. The \ac{LLM} prompt includes three in-context examples. When recovery is attempted, refinement is limited to three iterations. If the error persists, the random fallback strategy is applied. This setup allows for systematic comparison of the robustness of these recovery methods in the autoformalisation process. The generated warnings use three heuristics to identify semantic errors. First, checking for predicates and individuals only mentioned in the query to avoid an incomplete context. Second, ensuring that all predicates with the same identifier use the same number of parameters (checking for n-arity). Finally, avoiding similarly named predicates and individuals based on Levenshtein distance.

%% file: sections/evaluation.tex
\section{Evaluation}
We provide three sets of insights into this section, organised as \textit{findings (F*)}. We quantitatively study the effect of adversarial and counterfactual perturbations on the performance of informal reasoners and autoformalisation methods. Then, we dive deeper into method variants. Finally, 
we analyse the nature of formalisation errors made by the models.

\subsection{Robustness Analysis}
\paragraph{\textbf{\emph{F1: Noise perturbations have a stronger effect on autoformalisation methods than informal \ac{LLM} reasoners.}}}
Table~\ref{tab:distraction_k4_formalisation} shows that, on average, the accuracy of both direct and \ac{CoT} informal reasoning remains between $73\%$ and $74\%$ in the face of added noise. While the autoformalisation method performs similarly to informal reasoners on the original dataset, its performance decreases between $4\%$ and $11\%$. The accuracy drops especially with logical (L) and tautological (T) distractions, whose language formats trick the \ac{LLM} into formalising the noisy clauses. On the other hand, the linguistically complex and more natural sentences of encyclopedic distractions show a minor effect, suggesting that \acp{LLM} successfully avoids formalising the more complicated but semantically less similar noise sentences. 

\paragraph{\textbf{\emph{F2: All \ac{LLM}-based reasoning methods suffer a drop for counterfactual perturbations.}}}
Table~\ref{tab:distraction_k4_formalisation} shows that counterfactual statements cause a significant decrease in performance for both the informal reasoners and autoformalisation methods of between $12\%$ and $13\%$ on average. 
Moreover, this observation also holds for all tested models, i.e., none are robust towards counterfactual perturbations across every evaluated dimension. Even the strongest model, GPT 4o-mini, yields a performance of 63-68\%, which is relatively close to the random performance of 50\%. The high impact of counterfactual statements (the single ``not'' inserted) could be due to the inability of \acp{LLM} to overwrite prior knowledge with explicitly stated information or memorisation of the answers. We study the error sources further in §\ref{subsec:errors}.  

\begin{figure}[t]
    \centering
    \scriptsize
    \begin{tikzpicture}
        \begin{axis}[name=gpt,
            title={GPT-4o-mini},
            width=0.6\linewidth,
            height=0.6\linewidth,
            xlabel={\# Noise sentences},
            ylabel={Accuracy},
            xmin=-0.1, xmax=4.1,
            ymin=0.5, ymax=0.9,
            xtick={1,2,4},
            ytick={0.55, 0.6, 0.65, 0.75, 0.8, 0.85},
            title style={yshift=-0.6em},
            legend style={at={(1,-0.20)},
	           anchor=north,legend columns=-1},
            x label style={at={(axis description cs:1,-0.05)},anchor=north},
            y label style={at={(axis description cs:-0.15,0.5)},anchor=south},
            ymajorgrids=true,
            grid style=dashed,
        ]
            \addplot[color=blue, mark=square,]
                coordinates {
                (0,0.848076939582825)(1,0.823076903820038)(2,0.826923072338104)(4,0.821153819561005)
                };
            \addplot[color=red, mark=triangle,]
                coordinates {
                (0,0.848076939582825)(1,0.817307710647583)(2,0.801923096179962)(4,0.759615361690521)
                };
            \addplot[color=green, mark=diamond,] 
                coordinates {
                (0,0.848076939582825)(1,0.767307698726654)(2,0.769230782985687)(4,0.803846180438995)
                };
            \addplot[color=blue, mark=square*] 
                coordinates {
                (0,0.627777755260468)(1,0.622222244739533)(2,0.600000023841858)(4,0.633333325386047)
                };
            \addplot[color=red, mark=triangle*,] 
                coordinates {
                (0,0.627777755260468)(1,0.611111104488373)(2,0.611111104488373)(4,0.594444453716278)
                };
            \addplot[color=green, mark=diamond*,] 
                coordinates {
                (0,0.627777755260468)(1,0.572222232818604)(2,0.538888871669769)(4,0.555555582046509)
                };
                \legend{E,L,T,$\text{E}_C$, $\text{L}_C$ , $\text{T}_C$}
        \end{axis}

        \begin{axis}[name=llama, at={($(gpt.east)+(0.1cm,0)$)},anchor=west,
            title={Llama 3.1 70b},
            width=0.6\linewidth,
            height=0.6\linewidth,
            xmin=-0.1,, xmax=4.1,
            ymin=0.5, ymax=0.9,
            xtick={1,2,4},
            ytick={0.55, 0.6, 0.65, 0.75, 0.8, 0.85},
            title style={yshift=-0.6em},
            yticklabel=\empty,
            ymajorgrids=true,
            grid style=dashed,
        ]
            \addplot[color=blue, mark=square,]
                coordinates {
                (0,0.838461518287659)(1,0.817307710647583)(2,0.805769205093384)(4,0.817307710647583)
                };
            \addplot[color=red, mark=triangle,]
                coordinates {
                (0,0.838461518287659)(1,0.819230794906616)(2,0.803846180438995)(4,0.771153867244721)
                };
            \addplot[color=green, mark=diamond,]
                coordinates {
                (0,0.838461518287659)(1,0.803846180438995)(2,0.807692289352417)(4,0.805769205093384)
                };
            \addplot[color=blue, mark=square*]
                coordinates {
                (0,0.627777755260468)(1,0.622222244739533)(2,0.577777802944183)(4,0.594444453716278)
                };
            \addplot[color=red, mark=triangle*,]
                coordinates {
                (0,0.627777755260468)(1,0.583333313465118)(2,0.561111092567444)(4,0.577777802944183)
                };
            \addplot[color=green, mark=diamond*,]
                coordinates {
                (0,0.627777755260468)(1,0.627777755260468)(2,0.566666662693024)(4,0.577777802944183)
                };
        \end{axis}
    \end{tikzpicture}
    \caption{Influence of the number of noisy sentences for FOL.}
    \label{fig:length_distraction}
\end{figure}
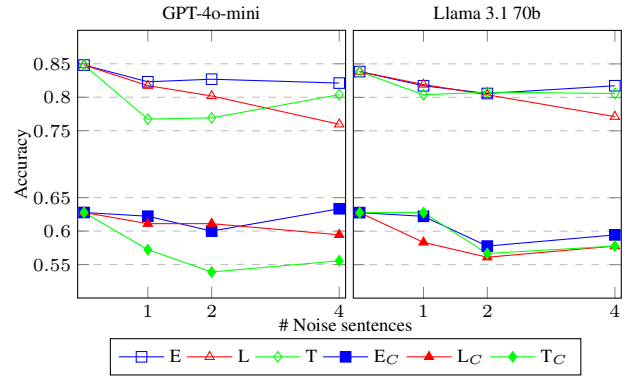

\noindent \paragraph{\textbf{\emph{F3: Introducing multiple noise sentences has an effect only for logical distractions.}}}
We show the impact of introducing between one and four sentences for the two top-performing autoformalisation models in Figure~\ref{fig:length_distraction}. The figure shows similar trends with and without counterfactual perturbations.
As additional logical distractions are introduced, the model performance consistently decreases. Tautological (T) distractions lead to a decline in accuracy with a single disruptive sentence, yet adding more noise does not worsen the outcome. 
The tautological corpus introduces truth constants for all sentences as a persistent unseen logical construct. Given that this leads only to a decrease for a single occurrence, we can assume that a model can consistently handle the same unseen logical construct. In contrast, the logical corpus increases the chance of adding text, requiring new, previously unseen reasoning constructs for each added sentence. The impact of encyclopedic noise remains negligible, generalising F1 to $k$ sentences. Similarly, counterfactual perturbations remain much more effective for all settings, generalising F2.

\begin{table}[t]
\small
\setlength{\modelspacing}{2pt}
\setlength{\tabcolsep}{1.7pt} % Default value: 6pt
\setlength{\belowrulesep}{4pt}
\begin{threeparttable}
    \centering
    \begin{tabular}{cc l r rrr @{\quad} rrrr}
\toprule
\multirow{2}{*}{} & \multirow{2}{*}{} & Reasoning & \multirow{2}{*}{O} & \multicolumn{3}{c}{Distraction} & \multicolumn{4}{c}{Counterfactual} \\
 & & Format & & E& L & T & $\text{O}_C$ & $\text{E}_C$& $\text{L}_C$ & $\text{T}_C$\\
\midrule
\multirow{6}{*}{\rotatebox{90}{Gemma-2}} & \multirow{3}{*}{\rotatebox{90}{9b}}
   & Informal (direct) & \textbf{0.78} & \textbf{0.80} & \textbf{0.79} & \textbf{0.77} & 0.58 & 0.52 & 0.50 & 0.59 \\
 & & Informal (CoT) & 0.72 & 0.78 & 0.73 & 0.76 & 0.61 & \textbf{0.57} & \textbf{0.60} & \textbf{0.66} \\
 & & Formal (FOL) & 0.62 & 0.58 & 0.52 & 0.53 & \textbf{0.63} & 0.52 & 0.46 & 0.46 \\[\modelspacing]
\cmidrule{2-11}
 & \multirow{3}{*}{\rotatebox{90}{27b}} 
   & Informal (direct) & 0.71 & 0.69 & \textbf{0.66} & \textbf{0.68} & 0.59 & 0.51 & 0.54 & 0.59 \\
 & & Informal (CoT) & 0.66 & 0.65 & 0.64 & 0.63 & 0.62 & 0.58 & \textbf{0.62} & \textbf{0.64} \\
 & & Formal (FOL) & \textbf{0.74} & \textbf{0.74} & 0.61 & 0.61 & \underline{\textbf{0.72}} & \underline{\textbf{0.67}} & 0.58 & 0.51 \\[\modelspacing]
\midrule
\multirow{6}{*}{\rotatebox{90}{Mistral}} & \multirow{3}{*}{\rotatebox{90}{7B}} 
   & Informal (direct) & 0.77 & \textbf{0.77} & 0.75 & \textbf{0.79} & \textbf{0.63} & \textbf{0.54} & \textbf{0.54} & \textbf{0.66} \\
 & & Informal (CoT) & \textbf{0.79} & 0.75 & \textbf{0.77} & 0.78 & 0.55 & 0.52 & \textbf{0.54} & 0.58 \\
 & & Formal (FOL) & 0.62 & 0.58 & 0.54 & 0.57 & 0.50 & \textbf{0.54} & 0.51 & 0.52 \\[\modelspacing]
\cmidrule{2-11}
 & \multirow{3}{*}{\rotatebox{90}{Small}} 
   & Informal (direct) & \textbf{0.77} & \textbf{0.76} & \textbf{0.76} & \textbf{0.75} & 0.61 & 0.51 & 0.56 & 0.59 \\
 & & Informal (CoT) & 0.72 & 0.72 & 0.72 & 0.71 & \textbf{0.62} & \textbf{0.59} & \textbf{0.62} & \textbf{0.68} \\
 & & Formal (FOL) & 0.68 & 0.59 & 0.53 & 0.64 & 0.54 & 0.55 & 0.49 & 0.51 \\[\modelspacing]
\midrule
\multirow{6}{*}{\rotatebox{90}{Llama-3.1}} & \multirow{3}{*}{\rotatebox{90}{8B}} 
   & Informal (direct) & 0.63 & 0.61 & 0.64 & 0.66 & 0.61 & \textbf{0.62} & 0.59 & 0.61 \\
 & & Informal (CoT) & 0.73 & \textbf{0.73} & \textbf{0.71} & \textbf{0.72} & \textbf{0.62} & 0.59 & \textbf{0.61} & \textbf{0.65} \\
 & & Formal (FOL) & \textbf{0.77} & 0.71 & 0.63 & 0.52 & 0.60 & 0.58 & 0.55 & 0.52 \\[\modelspacing]
\cmidrule{2-11}
 & \multirow{3}{*}{\rotatebox{90}{70B}} 
   & Informal (direct) & 0.77 & 0.74 & 0.74 & 0.73 & 0.62 & 0.53 & 0.56 & 0.64 \\
 & & Informal (CoT) & \textbf{0.78} & \textbf{0.75} & \textbf{0.76} & \textbf{0.76} & 0.64 & 0.61 & \textbf{0.66} & \underline{\textbf{0.73}} \\
 & & Formal (FOL) & 0.74 & 0.73 & 0.71 & 0.71 & \textbf{0.66} & \textbf{0.62} & 0.59 & 0.57 \\[\modelspacing]
 \midrule
\multirow{3}{*}{\rotatebox{90}{GPT}} & \multirow{3}{*}{\rotatebox{90}{4o-mini}} 
   & Informal (direct) & 0.78 & 0.77 & 0.79 & 0.79 & 0.64 & 0.61 & 0.61 & 0.63 \\
 & & Informal (CoT) & 0.80 & 0.80 & \underline{\textbf{0.81}} & \underline{\textbf{0.82}} & \textbf{0.68} & \textbf{0.63} & \underline{\textbf{0.68}} & \textbf{0.64} \\
 & & Formal (FOL) & \underline{\textbf{0.84}} & \underline{\textbf{0.82}} & 0.73 & 0.79 & 0.63 & 0.62 & 0.57 & 0.54 \\[\modelspacing]
 \midrule
\multicolumn{2}{c}{\multirow{3}{*}{\textbf{Avg}}} 
 & Informal (direct) & 0.74 & 0.73 & 0.73 & 0.73 & 0.61 & 0.55 & 0.56 & 0.62 \\
 & & Informal (CoT) & 0.74 & 0.74 & 0.73 & 0.74 & 0.62 & 0.58 & 0.62 & 0.65 \\
  & & Formal (FOL) & 0.72 & 0.68 &	0.61 & 0.62 & 0.61 & 0.59 & 0.54 & 0.52 \\
\bottomrule
\end{tabular}
\caption{Accuracies of informal and autoformalisation-based deductive reasoners. The best overall model per dataset is underlined; the best model version is marked in bold.}
\label{tab:distraction_k4_formalisation}
\end{threeparttable}
\end{table} 

\subsection{Impact of Method Design}
\paragraph{\textbf{\emph{F4: \ac{CoT} prompting is most impactful when both noise and counterfactual perturbations are applied.}}}
The accuracies for the individual \acp{LLM} in Table~\ref{tab:distraction_k4_formalisation} show that the impact of \ac{CoT} is negligible for noise-only datasets (first four columns). Meanwhile, the benefit from \ac{CoT} is most pronounced in the datasets that combine noise and counterfactual perturbations.
The better-performing informal prompting strategy for a model remains stable for all types of distractions. Still, the decline in performance due to counterfactuals leads to a less consistent preference for a specific prompting style.

\paragraph{\textbf{\emph{F5: The best-performing grammar differs per model and is unstable across data versions.}}}

The evaluation of different logical forms for formal \ac{LLM}-based reasoning in Table~\ref{tab:distraction_k4_logical_form} shows the preference of some models for specific syntactic formats.
Llama 3.1 70B has a considerable improvement of $12\%$ with TPTP syntax on the original set, while Llama 3.1 8B benefits from the R-FOL syntax. However, all grammars show a declining accuracy trend and increased syntax errors for noise perturbations, where the best grammar loses its advantage over the rest. 
When comparing the grammars on the counterfactual partitions, we observe that TPTP is consistently more robust than the standard first-order logic grammar. Here, GPT 4o-mini shows a reduction from $O$ to $O_C$ of $20\%$ for FOL and only $12\%$ for the TPTP grammar. Since this does not correlate with fewer syntax errors, the formalisation in TPTP prevents semantic errors for counterfactual premises. 
A positive reading of these results, especially the minor differences between FOL and R-FOL, is that autoformalisation \acp{LLM} can adapt to the grammar syntax prescribed in the prompt without further loss in performance.

\begin{table}[t]
\small
\setlength{\modelspacing}{2pt}
\setlength{\tabcolsep}{1.7pt} % Default value: 6pt
\setlength{\belowrulesep}{4pt}
\begin{threeparttable}
    \centering
    \begin{tabular}{cc l r rrr @{\quad} rrrr}
\toprule
\multirow{2}{*}{} & \multirow{2}{*}{} & Grammar & \multirow{2}{*}{O} & \multicolumn{3}{c}{Distraction} & \multicolumn{4}{c}{Counterfactual} \\
 & & Syntax & & E& L & T & $\text{O}_C$ & $\text{E}_C$& $\text{L}_C$ & $\text{T}_C$\\
\midrule
\multirow{6}{*}{\rotatebox{90}{Llama-3.1}} & \multirow{3}{*}{\rotatebox{90}{8B}} 
   & FOL & 0.77 & \textbf{0.71} & 0.61 & \textbf{0.53} & 0.58 & \textbf{0.55} & 0.52 & \textbf{0.56} \\
 & & R-FOL & \textbf{0.78} & 0.69 & \textbf{0.62} & \textbf{0.53} & 0.58 & \textbf{0.55} & \textbf{0.54} & 0.52 \\
 & & TPTP & 0.73 & 0.67 & 0.55 & 0.51 & \textbf{0.68} & 0.54 & 0.46 & 0.51 \\[\modelspacing]
\cmidrule{2-11}
 & \multirow{3}{*}{\rotatebox{90}{70B}} 
   & FOL & 0.76 & 0.73 & 0.71 & \textbf{0.72} & 0.67 & 0.57 & 0.63 & 0.56 \\
 & & R-FOL & 0.76 & 0.73 & 0.67 & 0.71 & 0.64 & 0.57 & 0.53 & 0.64 \\
 & & TPTP & \underline{\textbf{0.88}} & \underline{\textbf{0.84}} & \underline{\textbf{0.81}} & \textbf{0.72} & \underline{\textbf{0.81}} & \underline{\textbf{0.68}} & \underline{\textbf{0.67}} & \underline{\textbf{0.68}} \\[\modelspacing]
\midrule
\multirow{3}{*}{\rotatebox{90}{GPT}} & \multirow{3}{*}{\rotatebox{90}{4o-mini}} 
   & FOL & \textbf{0.84} & \textbf{0.82} & \textbf{0.72} & \underline{\textbf{0.78}} & 0.64 & \textbf{0.63} & \textbf{0.61} & 0.51 \\
 & & R-FOL & \textbf{0.84} & 0.77 & 0.70 & \underline{\textbf{0.78}} & \textbf{0.72} & 0.56 & 0.54 & \textbf{0.63} \\
 & & TPTP & 0.83 & \textbf{0.82} & 0.71 & 0.71 & 0.69 & \textbf{0.63} & 0.57 & 0.57 \\
\bottomrule
\end{tabular}
\caption{Accuracies of different formalisation grammars for autoformalisation.}
\label{tab:distraction_k4_logical_form}
\end{threeparttable}
\end{table} 

\paragraph{\textbf{\emph{F6: Feedback does not help \acp{LLM} self-correct to mitigate robustness issues.}}}
Table~\ref{tab:distraction_k4_feedback} shows the results with different error recovery mechanisms. The results indicate that no feedback strategy emerges as a winner in the various datasets. 
All feedback variants reduce syntax errors for noise perturbations, but given the lack of a consistent increase in accuracy, the corrected formalisations are still most likely to contain semantic errors. 
The type of feedback message only has a minor influence on correcting syntax errors, whereas Llama 3.1 70b and GPT 4o-mini correct slightly more syntax errors with specific error messages. This finding aligns with \cite{huang2023large}, who also found that \acp{LLM} cannot consistently self-correct their reasoning after receiving relevant feedback.

\begin{table}[t]
\small
\setlength{\modelspacing}{2pt}
\setlength{\tabcolsep}{1.7pt} % Default value: 6pt
\setlength{\belowrulesep}{4pt}
\begin{threeparttable}
    \centering
    \begin{tabular}{cc l r rrr @{\quad} rrrr}
\toprule
\multirow{2}{*}{} & \multirow{2}{*}{} & \multirow{2}{*}{Feedback} & \multirow{2}{*}{O} & \multicolumn{3}{c}{Distraction} & \multicolumn{4}{c}{Counterfactual} \\
 & & & & E& L & T & $\text{O}_C$ & $\text{E}_C$& $\text{L}_C$ & $\text{T}_C$\\
\midrule
\multirow{8}{*}{\rotatebox{90}{Llama-3.1}} & \multirow{4}{*}{\rotatebox{90}{8B}} 
   & No recovery & 0.77 & \textbf{0.72} & 0.62 & 0.53 & 0.59 & 0.58 & 0.56 & \textbf{0.56} \\
 & & Error type & \textbf{0.79} & 0.71 & 0.63 & \textbf{0.56} & \textbf{0.66} & 0.54 & 0.52 & 0.51 \\
 & & Error message & 0.78 & 0.71 & \textbf{0.67} & 0.55 & 0.59 & 0.53 & \underline{\textbf{0.64}} & 0.49 \\
 & & Warning & 0.74 & 0.66 & 0.58 & 0.55 & 0.55 & \textbf{0.60} & 0.49 & 0.49 \\[\modelspacing]
\cmidrule{2-11}
 & \multirow{4}{*}{\rotatebox{90}{70B}} 
   & No recovery & \textbf{0.77} & \textbf{0.72} & \textbf{0.73} & 0.71 & \textbf{0.64} & 0.59 & \textbf{0.61} & 0.56 \\
 & & Error type & 0.72 & 0.70 & 0.72 & \textbf{0.73} & 0.62 & 0.56 & 0.60 & 0.58 \\
 & & Error message & 0.71 & 0.70 & \textbf{0.73} & 0.71 & \textbf{0.64} & 0.59 & 0.54 & \underline{\textbf{0.64}} \\
 & & Warning & 0.69 & \textbf{0.72} & 0.72 & 0.72 & 0.62 & \underline{\textbf{0.65}} & \textbf{0.61} & 0.63 \\[\modelspacing]
\midrule
\multirow{4}{*}{\rotatebox{90}{GPT}} & \multirow{4}{*}{\rotatebox{90}{4o-mini}} 
   & No recovery & \underline{\textbf{0.84}} & \underline{\textbf{0.82}} & 0.73 & 0.79 & 0.64 & \textbf{0.62} & 0.56 & \textbf{0.56} \\
 & & Error type & 0.83 & 0.79 & 0.74 & 0.76 & 0.67 & 0.57 & 0.56 & \textbf{0.56} \\
 & & Error message & \underline{\textbf{0.84}} & 0.78 & \underline{\textbf{0.77}} & \underline{\textbf{0.80}} & 0.62 & 0.59 & 0.56 & \textbf{0.56} \\
 & & Warning & \underline{\textbf{0.84}} & 0.75 & 0.73 & 0.76 & \underline{\textbf{0.70}} & 0.61 & \textbf{0.61} & 0.55 \\
 \bottomrule
\end{tabular}
\caption{Accuracies of error recovery strategies.}
\label{tab:distraction_k4_feedback}
\end{threeparttable}
\end{table} 

\subsection{Error Analysis}
\label{subsec:errors}

\paragraph{\textbf{\emph{F7: Autoformalisation increases syntax errors for noise perturbations.}}}
The low performance for noise perturbations correlates with more syntax errors for all models and distraction categories (cf. execution rates in Table~\ref{tab:appendix_k4_formalisation_exec}). The three worst-performing models (both Mistral models, Gemma-2 9b) generate, at best, for $37\%$  and, at worst, for only $4\%$ of the samples, a valid logical form. 
In particular, these models show an initial low execution rate between $41\%$ and $53\%$. The error message feedback reveals issues with the template structure, including using incorrect keywords or adding conversational phrases. Noise perturbations emphasize this effect causing the increased syntax errors. 
The second common syntax errors are perturbation-related. The added noise triggers the \acp{LLM} to hallucinate new syntax rules, e.g., introducing undefined truth constants as part of tautological distractions or switching from a camelCase to snake\_case naming schema as part of logical distractions, which is likely caused by the increased difference between in-context examples and the test sample. 
The error analysis suggests that larger models are more robust towards noise perturbations. Gemma-2 9b and Llama3.1 8b produce more syntax errors than their larger counterparts.

As the increased syntax errors are mainly consistent for specific noise types, e.g., tautological noise creates undefined truth constants, it may be suboptimal to use a single, larger \ac{LLM} for a broad set of formalisation tasks. Instead a framework with specialized agents that address particular syntax errors could mitigate these errors more effectively. 
Additionally, the accuracy of syntactically valid samples is higher than the informal reasoning methods for most distractions (Table~\ref{tab:appendix_k4_formalisation_vacc}), motivating a hybrid strategy: use formal reasoning when a valid formalisation is produced, and otherwise rely on informal reasoning as a backup.  

\paragraph{\textbf{\emph{F8: Autoformalisation increases semantic errors for counterfactuals.}}}
Unlike the introduced noise, counterfactual perturbations do not lead to more syntax errors. The execution rate in Table~\ref{tab:appendix_k4_formalisation_exec} is stable or improves for counterfactuals. However, we see a drop in accuracy for the counterfactual column $\text{O}_C$ in Table~\ref{tab:distraction_k4_formalisation} and can conclude that the number of logical forms with semantic errors has to increase. This suggests that the introduced negation is not correctly formalised with respect to the semantics of these samples. 
Instead of genuine deductive reasoning capabilities informal and autoformalisation rely on prior knowledge.  

Devising principled mechanisms for verifying semantic validity (e.g., ensuring that predicates are defined before being used in a query) could reduce semantic errors for counterfactuals. 
Looking at the warnings generated by the feedback mechanism, for GPT 4o-mini, $161$ warning messages are generated on the unperturbed data pointing towards specific semantic errors. $54$ of these were fixed with a single iteration. Across all models, the most frequent warning is not considering predicates and individuals as part of the context.
This suggest that verification feedback is able to verify semantic validity and incorporating this feedback can mitigate semantic errors. 

%% file: sections/conclusion.tex
\section{Conclusion}
We presented the first study of the robustness of formal and informal \ac{LLM}-based deductive reasoning methods by introducing two types of perturbations: adversarial noise and counterfactual statements. These perturbations were used to examine the methodological aspects of \ac{LLM} reasoners based on their format, syntax, and feedback mechanism for error recovery. While adversarial noise only affects autoformalisation approaches, counterfactual statements remain a significant challenge for all variants of the tested method. While feedback strategies may lead to fewer syntax errors in autoformalisation methods, the refined formalisations tend to be semantically incorrect, failing to increase accuracy. We call on future work to devise more advanced mechanisms for detecting, reporting, and incorporating semantic errors. One possible mechanism to mitigate these issues could be the usage of models fine-tuned on error correction data. We also anticipate generalising the study in this paper to other logical deduction datasets and examining characteristics like the placement of adversarial noise and the effects of quantisation
to strengthen the generalizability of the reported insights.

%% file: sections/appendix.tex
\section{Experiments}
\label{app:experiments}

\begin{table}[!ht]
\small
\setlength{\modelspacing}{2pt}
\setlength{\tabcolsep}{1.7pt} % Default value: 6pt
\setlength{\belowrulesep}{4pt}
\begin{threeparttable}
    \centering
    \begin{tabular}{cc l r rrr @{\quad} rrrr}
\toprule
\multirow{2}{*}{} & \multirow{2}{*}{} & \multirow{2}{*}{Formalisation} & \multirow{2}{*}{O} & \multicolumn{3}{c}{Distraction} & \multicolumn{4}{c}{Counterfactual} \\
 & & & & E& L & T & $\text{O}_C$ & $\text{E}_C$& $\text{L}_C$ & $\text{T}_C$\\
\midrule
\multirow{6}{*}{\rotatebox{90}{Gemma-2}} & \multirow{3}{*}{\rotatebox{90}{9b}}
   & Informal (direct) & 1.00 & 1.00 & 1.00 & 1.00 & 1.00 & 1.00 & 1.00 & 1.00 \\
 & & Informal (CoT) & 0.99 & 0.99 & 0.98 & 0.98 & 0.99 & 0.99 & 0.99 & 0.99 \\
 & & Formal (FOL) & 0.41 & 0.17 & 0.04 & 0.09 & 0.49 & 0.15 & 0.07 & 0.04 \\
\cmidrule{2-11}
 & \multirow{3}{*}{\rotatebox{90}{27b}} 
   & Informal (direct) & 1.00 & 1.00 & 1.00 & 1.00 & 1.00 & 1.00 & 1.00 & 1.00 \\
 & & Informal (CoT) & 1.00 & 1.00 & 1.00 & 1.00 & 1.00 & 0.99 & 1.00 & 0.99 \\
 & & Formal (FOL) & 0.71 & 0.72 & 0.37 & 0.24 & 0.74 & 0.71 & 0.34 & 0.17 \\
\midrule
\multirow{6}{*}{\rotatebox{90}{Mistral}} & \multirow{3}{*}{\rotatebox{90}{7B}} 
     & Informal (direct) & 0.99 & 0.99 & 0.99 & 0.99 & 0.98 & 0.97 & 0.98 & 0.97 \\
 & & Informal (CoT) & 0.96 & 0.95 & 0.96 & 0.95 & 0.93 & 0.96 & 0.98 & 0.93 \\
 & & Formal (FOL) & 0.51 & 0.37 & 0.23 & 0.33 & 0.61 & 0.38 & 0.24 & 0.29 \\
\cmidrule{2-11}
 & \multirow{3}{*}{\rotatebox{90}{Small}} 
    & Informal (direct) & 1.00 & 1.00 & 1.00 & 1.00 & 1.00 & 1.00 & 1.00 & 1.00 \\
 & & Informal (CoT) & 1.00 & 1.00 & 1.00 & 1.00 & 1.00 & 0.99 & 1.00 & 1.00 \\
 & & Formal (FOL) & 0.53 & 0.34 & 0.12 & 0.36 & 0.62 & 0.36 & 0.14 & 0.36 \\
\midrule
\multirow{6}{*}{\rotatebox{90}{Llama-3.1}} & \multirow{3}{*}{\rotatebox{90}{8B}} 
      & Informal (direct) & 1.00 & 1.00 & 1.00 & 1.00 & 1.00 & 1.00 & 1.00 & 1.00 \\
 & & Informal (CoT) & 1.00 & 1.00 & 0.98 & 0.98 & 0.98 & 1.00 & 0.99 & 0.93 \\
 & & Formal (FOL) & 0.83 & 0.70 & 0.47 & 0.28 & 0.86 & 0.70 & 0.40 & 0.37 \\
\cmidrule{2-11}
 & \multirow{3}{*}{\rotatebox{90}{70B}} 
    & Informal (direct) & 1.00 & 1.00 & 1.00 & 1.00 & 1.00 & 1.00 & 1.00 & 1.00 \\
 & & Informal (CoT) & 0.99 & 1.00 & 1.00 & 0.99 & 1.00 & 0.99 & 0.99 & 1.00 \\
 & & Formal (FOL) & 0.78 & 0.78 & 0.70 & 0.66 & 0.88 & 0.84 & 0.73 & 0.66 \\
\midrule
\multirow{3}{*}{\rotatebox{90}{GPT}} & \multirow{3}{*}{\rotatebox{90}{4o-mini}} 
    & Informal (direct) & 1.00 & 1.00 & 1.00 & 1.00 & 1.00 & 1.00 & 1.00 & 1.00 \\
 & & Informal (CoT) & 1.00 & 1.00 & 1.00 & 1.00 & 1.00 & 1.00 & 1.00 & 1.00 \\
 & & Formal (FOL) & 0.98 & 0.96 & 0.74 & 0.87 & 0.99 & 0.93 & 0.69 & 0.85 \\
\bottomrule
\end{tabular}
\caption{Comparison of execution rate of informal and autoformalisation-based LLM-based deductive reasoners.}
\label{tab:appendix_k4_formalisation_exec}
\end{threeparttable}
\end{table} 

\begin{table}[!ht]
\small
\setlength{\modelspacing}{2pt}
\setlength{\tabcolsep}{1.7pt} % Default value: 6pt
\setlength{\belowrulesep}{4pt}
\begin{threeparttable}
    \centering
    \begin{tabular}{cc l r rrr @{\quad} rrrr}
\toprule
\multirow{2}{*}{} & \multirow{2}{*}{} & \multirow{2}{*}{Formalisation} & \multirow{2}{*}{O} & \multicolumn{3}{c}{Distraction} & \multicolumn{4}{c}{Counterfactual} \\
 & & & & E& L & T & $\text{O}_C$ & $\text{E}_C$& $\text{L}_C$ & $\text{T}_C$\\
\midrule
\multirow{6}{*}{\rotatebox{90}{Gemma-2}} & \multirow{3}{*}{\rotatebox{90}{9b}}
   & Informal (direct) & 0.78 & 0.80 & \textbf{0.79} & 0.77 & 0.58 & 0.52 & 0.50 & 0.59 \\
 & & Informal (CoT) & 0.72 & 0.78 & 0.73 & 0.75 & 0.61 & \textbf{0.58} & 0.60 & \textbf{0.66} \\
 & & Formal (FOL) & \textbf{0.81} & \underline{\textbf{0.83}} & 0.70 & \underline{\textbf{0.91}} & \textbf{0.72} & \textbf{0.58} & \textbf{0.73} & 0.57 \\
\cmidrule{2-11}
 & \multirow{3}{*}{\rotatebox{90}{27b}} 
   & Informal (direct) & 0.71 & 0.69 & 0.66 & 0.68 & 0.59 & 0.51 & 0.54 & 0.59 \\
 & & Informal (CoT) & 0.66 & 0.65 & 0.64 & 0.63 & 0.62 & 0.58 & 0.62 & \textbf{0.64} \\
 & & Formal (FOL) & \underline{\textbf{0.85}} & \textbf{0.82} & \textbf{0.78} & \textbf{0.83} & \underline{\textbf{0.75}} & \underline{\textbf{0.71}} & \underline{\textbf{0.74}} & 0.54 \\
\midrule
\multirow{6}{*}{\rotatebox{90}{Mistral}} & \multirow{3}{*}{\rotatebox{90}{7B}} 
     & Informal (direct) & 0.77 & \textbf{0.78} & 0.75 & 0.79 & \textbf{0.63} & \textbf{0.54} & 0.54 & \textbf{0.65} \\
 & & Informal (CoT) & \textbf{0.80} & 0.77 & \textbf{0.78} & \textbf{0.80} & 0.57 & \textbf{0.54} & 0.55 & 0.58 \\
 & & Formal (FOL) & 0.70 & 0.76 & 0.69 & 0.66 & 0.48 & 0.38 & \textbf{0.58} & 0.46 \\
\cmidrule{2-11}
 & \multirow{3}{*}{\rotatebox{90}{Small}} 
    & Informal (direct) & 0.77 & 0.76 & 0.76 & 0.75 & 0.61 & 0.51 & 0.56 & 0.59 \\
 & & Informal (CoT) & 0.72 & 0.72 & 0.72 & 0.71 & \textbf{0.62} & 0.60 & \textbf{0.62} & 0.68 \\
 & & Formal (FOL) & \textbf{0.82} & \textbf{0.77} & \underline{\textbf{0.86}} & \textbf{0.81} & 0.58 & \textbf{0.63} & 0.52 & \textbf{0.69} \\
\midrule
\multirow{6}{*}{\rotatebox{90}{Llama-3.1}} & \multirow{3}{*}{\rotatebox{90}{8B}} 
   & Informal (direct) & 0.63 & 0.61 & 0.64 & 0.66 & 0.61 & \textbf{0.62} & 0.59 & 0.61 \\
 & & Informal (CoT) & 0.73 & 0.74 & 0.72 & \textbf{0.72} & \textbf{0.62} & 0.59 & \textbf{0.61} & \textbf{0.65} \\
 & & Formal (FOL) & \textbf{0.82} & \textbf{0.80} & \textbf{0.78} & 0.65 & 0.61 & 0.60 & 0.55 & 0.58 \\
\cmidrule{2-11}
 & \multirow{3}{*}{\rotatebox{90}{70B}} 
    & Informal (direct) & 0.77 & 0.74 & 0.74 & 0.73 & 0.62 & 0.53 & 0.56 & 0.64 \\
 & & Informal (CoT) & 0.78 & 0.75 & 0.76 & 0.76 & 0.64 & 0.60 & \textbf{0.66} & \underline{\textbf{0.73}} \\
 & & Formal (FOL) & \textbf{0.82} & \textbf{0.80} & \textbf{0.80} & \textbf{0.83} & \textbf{0.66} & \textbf{0.63} & 0.64 & 0.61 \\
\midrule
\multirow{3}{*}{\rotatebox{90}{GPT}} & \multirow{3}{*}{\rotatebox{90}{4o-mini}} 
    & Informal (direct) & 0.78 & 0.77 & 0.79 & 0.79 & 0.64 & 0.61 & 0.61 & 0.63 \\
 & & Informal (CoT) & 0.80 & 0.80 & \textbf{0.81} & \textbf{0.82} & \textbf{0.68} & 0.63 & \textbf{0.68} & \textbf{0.64} \\
 & & Formal (FOL) & \underline{\textbf{0.85}} & \underline{\textbf{0.83}} & 0.80 & \textbf{0.82} & 0.64 & \textbf{0.64} & 0.63 & 0.57 \\
\bottomrule
\end{tabular}
\caption{Comparison of valid accuracy of informal and autoformalisation-based LLM-based deductive reasoners. }
\label{tab:appendix_k4_formalisation_vacc}
\end{threeparttable}
\end{table} 

\begin{table}[!ht]
\small
\setlength{\modelspacing}{2pt}
\setlength{\tabcolsep}{1.7pt} % Default value: 6pt
\setlength{\belowrulesep}{4pt}
\begin{threeparttable}
    \centering
    \begin{tabular}{cc l r rrr @{\quad} rrrr}
\toprule
\multirow{2}{*}{} & \multirow{2}{*}{} & Grammar & \multirow{2}{*}{O} & \multicolumn{3}{c}{Distraction} & \multicolumn{4}{c}{Counterfactual} \\
 & & Syntax & & E& L & T & $\text{O}_C$ & $\text{E}_C$& $\text{L}_C$ & $\text{T}_C$\\
\midrule
\multirow{6}{*}{\rotatebox{90}{Gemma-2}} & \multirow{3}{*}{\rotatebox{90}{9b}}
   & FOL & 0.63 & 0.55 & 0.53 & 0.56 & 0.55 & 0.52 & 0.49 & 0.52 \\
 & & R-FOL & 0.71 & 0.55 & 0.52 & 0.55 & 0.55 & 0.52 & 0.54 & 0.54 \\
 & & TPTP & 0.68 & 0.53 & 0.52 & 0.52 & 0.68 & 0.54 & 0.52 & 0.49 \\
\cmidrule{2-11}
 & \multirow{3}{*}{\rotatebox{90}{27b}} 
   & FOL & 0.73 & 0.73 & 0.59 & 0.59 & 0.70 & 0.62 & 0.56 & 0.55 \\
 & & R-FOL & 0.77 & 0.69 & 0.67 & 0.64 & 0.66 & 0.63 & 0.55 & 0.56 \\
 & & TPTP & 0.80 & 0.76 & 0.59 & 0.53 & 0.74 & 0.69 & 0.66 & 0.49 \\
\midrule
\multirow{6}{*}{\rotatebox{90}{Mistral}} & \multirow{3}{*}{\rotatebox{90}{7B}} 
   & FOL & 0.59 & 0.59 & 0.58 & 0.53 & 0.49 & 0.46 & 0.57 & 0.43 \\
 & & R-FOL & 0.60 & 0.60 & 0.52 & 0.58 & 0.49 & 0.49 & 0.49 & 0.54 \\
 & & TPTP & 0.65 & 0.61 & 0.51 & 0.50 & 0.53 & 0.46 & 0.47 & 0.48 \\
\cmidrule{2-11}
 & \multirow{3}{*}{\rotatebox{90}{Small}} 
   & FOL & 0.64 & 0.62 & 0.52 & 0.64 & 0.51 & 0.58 & 0.53 & 0.51 \\
 & & R-FOL & 0.71 & 0.64 & 0.53 & 0.60 & 0.52 & 0.57 & 0.54 & 0.56 \\
 & & TPTP & 0.66 & 0.58 & 0.54 & 0.48 & 0.59 & 0.52 & 0.49 & 0.50 \\
\bottomrule
\end{tabular}
\caption{Comparison of accuracies between different formalisation grammar syntaxes of autoformalisation.  }
\label{tab:appendix_k4_lf_full}
\end{threeparttable}
\end{table} 

\begin{table}[!ht]
\small
\setlength{\modelspacing}{2pt}
\setlength{\tabcolsep}{1.7pt} % Default value: 6pt
\setlength{\belowrulesep}{4pt}
\begin{threeparttable}
    \centering
    \begin{tabular}{cc l r rrr @{\quad} rrrr}
\toprule
\multirow{2}{*}{} & \multirow{2}{*}{} & Grammar & \multirow{2}{*}{O} & \multicolumn{3}{c}{Distraction} & \multicolumn{4}{c}{Counterfactual} \\
 & & Syntax & & E& L & T & $\text{O}_C$ & $\text{E}_C$& $\text{L}_C$ & $\text{T}_C$\\
\midrule
\multirow{6}{*}{\rotatebox{90}{Gemma-2}} & \multirow{3}{*}{\rotatebox{90}{9b}}
   & FOL & 0.41 & 0.17 & 0.04 & 0.09 & 0.49 & 0.15 & 0.07 & 0.04 \\
 & & R-FOL & \textbf{0.63} & \textbf{0.27} & \textbf{0.13} & \textbf{0.17} & \textbf{0.64} & \textbf{0.33} & \textbf{0.14} & \textbf{0.14} \\
 & & TPTP & 0.44 & 0.13 & 0.04 & 0.01 & 0.47 & 0.17 & 0.04 & 0.01 \\
\cmidrule{2-11}
 & \multirow{3}{*}{\rotatebox{90}{27b}} 
   & FOL & 0.71 & 0.72 & 0.37 & 0.24 & 0.74 & 0.71 & 0.34 & 0.17 \\
 & & R-FOL & \textbf{0.83} & 0.78 & \textbf{0.50} & \textbf{0.44} & \textbf{0.78} & 0.77 & \textbf{0.53} & \textbf{0.32} \\
 & & TPTP & 0.82 & \textbf{0.82} & 0.41 & 0.08 & 0.76 & \textbf{0.84} & 0.39 & 0.01 \\
\midrule
\multirow{6}{*}{\rotatebox{90}{Mistral}} & \multirow{3}{*}{\rotatebox{90}{7B}} 
   & FOL & 0.51 & 0.37 & 0.23 & \textbf{0.33} & 0.61 & 0.38 & 0.24 & 0.29 \\
 & & R-FOL & \textbf{0.54} & \textbf{0.45} & \textbf{0.24} & \textbf{0.33} & \textbf{0.68} & \textbf{0.39} & \textbf{0.25} & \textbf{0.32} \\
 & & TPTP & 0.45 & 0.31 & 0.08 & 0.10 & 0.50 & 0.26 & 0.07 & 0.07 \\
\cmidrule{2-11}
 & \multirow{3}{*}{\rotatebox{90}{Small}} 
   & FOL & 0.53 & 0.34 & 0.12 & 0.36 & 0.62 & 0.36 & 0.14 & 0.36 \\
 & & R-FOL & \textbf{0.68} & \textbf{0.41} & \textbf{0.17} & \textbf{0.38} & \textbf{0.69} & \textbf{0.38} & \textbf{0.18} & \textbf{0.38} \\
 & & TPTP & 0.45 & 0.25 & 0.05 & 0.03 & 0.41 & 0.28 & 0.06 & 0.02 \\
\midrule
\multirow{6}{*}{\rotatebox{90}{Llama-3.1}} & \multirow{3}{*}{\rotatebox{90}{8B}} 
   & FOL & \textbf{0.83} & \textbf{0.70} & \textbf{0.47} & \textbf{0.28} & \textbf{0.86} & \textbf{0.70} & 0.40 & \textbf{0.37} \\
 & & R-FOL & 0.81 & 0.65 & 0.38 & 0.25 & 0.84 & 0.64 & \textbf{0.43} & 0.31 \\
 & & TPTP & 0.75 & 0.49 & 0.21 & 0.01 & 0.74 & 0.47 & 0.20 & 0.02 \\
\cmidrule{2-11}
 & \multirow{3}{*}{\rotatebox{90}{70B}} 
   & FOL & 0.78 & 0.78 & 0.70 & 0.66 & 0.88 & 0.84 & 0.73 & 0.66 \\
 & & R-FOL & 0.80 & 0.81 & 0.65 & \textbf{0.76} & \textbf{0.93} & \textbf{0.89} & 0.67 & \textbf{0.72} \\
 & & TPTP & \textbf{0.92} & \textbf{0.93} & \underline{\textbf{0.80}} & 0.58 & \textbf{0.93} & \textbf{0.89} & \underline{\textbf{0.78}} & 0.51 \\
\midrule
\multirow{3}{*}{\rotatebox{90}{GPT}} & \multirow{3}{*}{\rotatebox{90}{4o-mini}} 
   & FOL & \underline{\textbf{0.98}} & \underline{\textbf{0.96}} & \textbf{0.74} & \underline{\textbf{0.87}} & \underline{\textbf{0.99}} & 0.93 & \textbf{0.69} & 0.85 \\
 & & R-FOL & 0.96 & 0.91 & 0.68 & 0.83 & 0.95 & \underline{\textbf{0.96}} & 0.62 & \underline{\textbf{0.87}} \\
 & & TPTP & 0.93 & 0.91 & 0.52 & 0.62 & 0.91 & 0.92 & 0.49 & 0.47 \\
\bottomrule
\end{tabular}
\caption{Comparison of execution rate between different formalisation grammar syntaxes of autoformalisation.}
\label{tab:appendix_k4_lf_exec}
\end{threeparttable}
\end{table} 

\begin{table}[!ht]
\small
\setlength{\modelspacing}{2pt}
\setlength{\tabcolsep}{1.7pt} % Default value: 6pt
\setlength{\belowrulesep}{4pt}
\begin{threeparttable}
    \centering
    \begin{tabular}{cc l r rrr @{\quad} rrrr}
\toprule
\multirow{2}{*}{} & \multirow{2}{*}{} & Grammar & \multirow{2}{*}{O} & \multicolumn{3}{c}{Distraction} & \multicolumn{4}{c}{Counterfactual} \\
 & & Syntax & & E& L & T & $\text{O}_C$ & $\text{E}_C$& $\text{L}_C$ & $\text{T}_C$\\
\midrule
\multirow{6}{*}{\rotatebox{90}{Gemma-2}} & \multirow{3}{*}{\rotatebox{90}{9b}}
   & FOL & 0.81 & \textbf{0.83} & 0.70 & \underline{\textbf{0.91}} & 0.72 & 0.58 & \textbf{0.73} & 0.57 \\
 & & R-FOL & 0.82 & 0.68 & 0.65 & 0.74 & 0.60 & 0.58 & 0.55 & 0.57 \\
 & & TPTP & \underline{\textbf{0.92}} & 0.75 & \textbf{0.74} & 0.67 & \textbf{0.79} & \textbf{0.63} & 0.57 & \underline{\textbf{1.00}} \\
\cmidrule{2-11}
 & \multirow{3}{*}{\rotatebox{90}{27b}} 
   & FOL & 0.85 & 0.82 & 0.78 & 0.83 & 0.75 & \underline{\textbf{0.71}} & 0.74 & 0.54 \\
 & & R-FOL & 0.82 & 0.75 & \textbf{0.81} & 0.81 & 0.70 & 0.66 & 0.65 & 0.65 \\
 & & TPTP & \textbf{0.87} & \textbf{0.83} & 0.79 & \textbf{0.88} & \textbf{0.79} & 0.69 & \underline{\textbf{0.76}} & \underline{\textbf{1.00}} \\
\midrule
\multirow{6}{*}{\rotatebox{90}{Mistral}} & \multirow{3}{*}{\rotatebox{90}{7B}} 
   & FOL & 0.70 & 0.76 & 0.69 & 0.66 & 0.48 & 0.38 & \textbf{0.58} & 0.46 \\
 & & R-FOL & 0.66 & 0.73 & 0.66 & 0.72 & 0.44 & 0.41 & 0.52 & 0.54 \\
 & & TPTP & \textbf{0.83} & \textbf{0.79} & \textbf{0.75} & \textbf{0.80} & \textbf{0.62} & \textbf{0.46} & 0.36 & \textbf{0.55} \\
\cmidrule{2-11}
 & \multirow{3}{*}{\rotatebox{90}{Small}} 
   & FOL & 0.82 & 0.77 & \textbf{0.86} & 0.81 & 0.58 & \textbf{0.63} & 0.52 & 0.69 \\
 & & R-FOL & 0.81 & 0.73 & 0.64 & 0.79 & 0.55 & 0.62 & \textbf{0.55} & 0.70 \\
 & & TPTP & \textbf{0.85} & \textbf{0.81} & 0.81 & \textbf{0.82} & \textbf{0.70} & 0.61 & 0.22 & \underline{\textbf{1.00}} \\
\midrule
\multirow{6}{*}{\rotatebox{90}{Llama-3.1}} & \multirow{3}{*}{\rotatebox{90}{8B}} 
   & FOL & 0.82 & \textbf{0.80} & 0.78 & 0.65 & 0.61 & 0.60 & 0.55 & 0.58 \\
 & & R-FOL & \textbf{0.84} & 0.77 & \textbf{0.80} & \textbf{0.74} & 0.62 & 0.58 & \textbf{0.59} & 0.56 \\
 & & TPTP & 0.82 & \textbf{0.80} & 0.72 & 0.67 & \textbf{0.76} & \textbf{0.68} & 0.50 & \textbf{0.67} \\
\cmidrule{2-11}
 & \multirow{3}{*}{\rotatebox{90}{70B}} 
   & FOL & 0.82 & 0.80 & 0.80 & 0.83 & 0.66 & 0.63 & 0.64 & 0.61 \\
 & & R-FOL & 0.81 & 0.77 & 0.77 & 0.77 & 0.67 & 0.61 & 0.58 & 0.64 \\
 & & TPTP & \textbf{0.90} & \underline{\textbf{0.88}} & \underline{\textbf{0.90}} & \textbf{0.90} & \underline{\textbf{0.84}} & \textbf{0.70} & \textbf{0.67} & \textbf{0.78} \\
\midrule
\multirow{3}{*}{\rotatebox{90}{GPT}} & \multirow{3}{*}{\rotatebox{90}{4o-mini}} 
   & FOL & 0.85 & 0.83 & 0.80 & 0.82 & 0.64 & 0.64 & 0.63 & 0.57 \\
 & & R-FOL & \textbf{0.86} & 0.80 & 0.80 & 0.83 & \textbf{0.72} & 0.56 & 0.59 & \textbf{0.63} \\
 & & TPTP & 0.85 & \textbf{0.85} & \textbf{0.85} & \textbf{0.85} & \textbf{0.72} & \textbf{0.66} & \textbf{0.70} & 0.61 \\
\bottomrule
\end{tabular}
\caption{Comparison of valid accuracies between different formalisation grammar syntaxes of autoformalisation. }
\label{tab:appendix_k4_lf_vacc}
\end{threeparttable}
\end{table} 

\begin{table}[!ht]
\small
\setlength{\modelspacing}{2pt}
\setlength{\tabcolsep}{1.7pt} % Default value: 6pt
\setlength{\belowrulesep}{4pt}
\begin{threeparttable}
    \centering
    \begin{tabular}{cc l r rrr @{\quad} rrrr}
\toprule
\multirow{2}{*}{} & \multirow{2}{*}{} & \multirow{2}{*}{Feedback} & \multirow{2}{*}{O} & \multicolumn{3}{c}{Distraction} & \multicolumn{4}{c}{Counterfactual} \\
 & & & & E& L & T & $\text{O}_C$ & $\text{E}_C$& $\text{L}_C$ & $\text{T}_C$\\
\midrule
\multirow{6}{*}{\rotatebox{90}{Gemma-2}} & \multirow{3}{*}{\rotatebox{90}{9b}}
   & No recovery & 0.66 & 0.58 & 0.52 & 0.58 & 0.58 & 0.46 & 0.49 & 0.53 \\
 & & Error type & 0.65 & 0.57 & 0.50 & 0.53 & 0.59 & 0.50 & 0.49 & 0.52 \\
 & & Error message & 0.62 & 0.55 & 0.53 & 0.55 & 0.58 & 0.56 & 0.50 & 0.65 \\
\cmidrule{2-11}
 & \multirow{3}{*}{\rotatebox{90}{27b}} 
   & No recovery & 0.74 & 0.74 & 0.60 & 0.59 & 0.67 & 0.64 & 0.53 & 0.49 \\
 & & Error type & 0.77 & 0.75 & 0.62 & 0.66 & 0.70 & 0.62 & 0.54 & 0.54 \\
 & & Error message & 0.78 & 0.73 & 0.64 & 0.62 & 0.70 & 0.68 & 0.58 & 0.51 \\
\midrule
\multirow{6}{*}{\rotatebox{90}{Mistral}} & \multirow{3}{*}{\rotatebox{90}{7B}} 
   & No recovery & 0.61 & 0.58 & 0.54 & 0.54 & 0.51 & 0.48 & 0.56 & 0.46 \\
 & & Error type & 0.65 & 0.57 & 0.56 & 0.61 & 0.47 & 0.49 & 0.51 & 0.48 \\
 & & Error message & 0.61 & 0.59 & 0.52 & 0.55 & 0.52 & 0.46 & 0.52 & 0.56 \\
\cmidrule{2-11}
 & \multirow{3}{*}{\rotatebox{90}{Small}} 
   & No recovery & 0.68 & 0.57 & 0.51 & 0.61 & 0.56 & 0.54 & 0.52 & 0.59 \\
 & & Error type & 0.70 & 0.59 & 0.58 & 0.66 & 0.54 & 0.61 & 0.62 & 0.58 \\
 & & Error message & 0.72 & 0.65 & 0.59 & 0.66 & 0.59 & 0.43 & 0.56 & 0.60 \\
\bottomrule
\end{tabular}
\caption{Comparison between accuracy of error recovery strategies. }
\label{tab:appendix_k4_feedback_full}
\end{threeparttable}
\end{table}

\begin{table}[!ht]
\small
\setlength{\modelspacing}{2pt}
\setlength{\tabcolsep}{1.7pt} % Default value: 6pt
\setlength{\belowrulesep}{4pt}
\begin{threeparttable}
    \centering
    \begin{tabular}{cc l r rrr @{\quad} rrrr}
\toprule
\multirow{2}{*}{} & \multirow{2}{*}{} & \multirow{2}{*}{Feedback} & \multirow{2}{*}{O} & \multicolumn{3}{c}{Distraction} & \multicolumn{4}{c}{Counterfactual} \\
 & & & & E& L & T & $\text{O}_C$ & $\text{E}_C$& $\text{L}_C$ & $\text{T}_C$\\
\midrule
\multirow{6}{*}{\rotatebox{90}{Gemma-2}} & \multirow{3}{*}{\rotatebox{90}{9b}}
   & No recovery & 0.41 & 0.17 & 0.04 & 0.09 & 0.49 & 0.15 & 0.07 & 0.04 \\
 & & Error type & \textbf{0.52} & 0.23 & 0.11 & \textbf{0.16} & \textbf{0.57} & 0.23 & 0.10 & \textbf{0.17} \\
 & & Error message & 0.51 & \textbf{0.26} & \textbf{0.12} & 0.14 & \textbf{0.57} & \textbf{0.26} & \textbf{0.11} & \textbf{0.17} \\
\cmidrule{2-11}
 & \multirow{3}{*}{\rotatebox{90}{27b}} 
   & No recovery & 0.71 & 0.72 & 0.37 & 0.24 & 0.74 & 0.71 & 0.34 & 0.17 \\
 & & Error type & 0.77 & 0.80 & 0.49 & \textbf{0.53} & 0.81 & 0.76 & 0.41 & \textbf{0.43} \\
 & & Error message & \textbf{0.88} & \textbf{0.83} & \textbf{0.54} & 0.50 & \textbf{0.84} & \textbf{0.85} & \textbf{0.52} & 0.33 \\
\midrule
\multirow{6}{*}{\rotatebox{90}{Mistral}} & \multirow{3}{*}{\rotatebox{90}{7B}} 
   & No recovery & 0.51 & 0.37 & 0.23 & 0.33 & 0.61 & 0.38 & 0.24 & 0.29 \\
 & & Error type & \textbf{0.66} & \textbf{0.54} & 0.30 & \textbf{0.47} & \textbf{0.68} & \textbf{0.57} & 0.26 & \textbf{0.47} \\
 & & Error message & 0.63 & 0.49 & \textbf{0.37} & 0.40 & 0.67 & 0.54 & \textbf{0.30} & 0.36 \\
\cmidrule{2-11}
 & \multirow{3}{*}{\rotatebox{90}{Small}} 
   & No recovery & 0.53 & 0.34 & 0.12 & 0.36 & 0.62 & 0.36 & 0.14 & 0.36 \\
 & & Error type & 0.64 & 0.41 & 0.25 & 0.51 & 0.62 & \textbf{0.52} & 0.22 & 0.46 \\
 & & Error message & \textbf{0.69} & \textbf{0.49} & \textbf{0.26} & \textbf{0.53} & \textbf{0.70} & 0.46 & \textbf{0.28} & \textbf{0.49} \\
\midrule
\multirow{8}{*}{\rotatebox{90}{Llama-3.1}} & \multirow{4}{*}{\rotatebox{90}{8B}} 
   & No recovery & 0.83 & 0.70 & 0.47 & 0.28 & 0.86 & 0.70 & 0.40 & 0.37 \\
 & & Error type & \textbf{0.93} & \textbf{0.82} & \textbf{0.55} & 0.34 & 0.91 & \textbf{0.79} & 0.52 & \textbf{0.43} \\
 & & Error message & \textbf{0.93} & \textbf{0.82} & 0.52 & 0.33 & \textbf{0.92} & 0.77 & \textbf{0.55} & 0.42 \\
  & & Warning & 0.86 & 0.74 & 0.51 & \textbf{0.35} & 0.87 & 0.76 & 0.47 & 0.38 \\
\cmidrule{2-11}
 & \multirow{4}{*}{\rotatebox{90}{70B}} 
   & No recovery & 0.78 & 0.78 & 0.70 & 0.66 & 0.88 & 0.84 & 0.73 & 0.66 \\
 & & Error type & 0.92 & 0.94 & 0.84 & \textbf{0.87} & 0.94 & 0.93 & 0.84 & 0.80 \\
 & & Error message & \textbf{0.96} & \textbf{0.97} & \underline{\textbf{0.88}} & \textbf{0.87} & \textbf{0.95} & \textbf{0.96} & 0.88 & \textbf{0.84} \\
  & & Warning & 0.94 & 0.96 & 0.87 & 0.84 & 0.92 & \textbf{0.96} & \underline{\textbf{0.89}} & 0.81 \\
\midrule
\multirow{4}{*}{\rotatebox{90}{GPT}} & \multirow{4}{*}{\rotatebox{90}{4o-mini}} 
   & No recovery & \underline{\textbf{0.98}} & 0.96 & 0.74 & 0.87 & \underline{\textbf{0.99}} & 0.93 & 0.69 & 0.85 \\
 & & Error type & \underline{\textbf{0.98}} & \underline{\textbf{0.98}} & 0.79 & 0.88 & 0.96 & \underline{\textbf{0.99}} & 0.73 & \underline{\textbf{0.88}} \\
 & & Error message & \underline{\textbf{0.98}} & 0.97 & \textbf{0.84} & \underline{\textbf{0.89}} & 0.98 & \underline{\textbf{0.99}} & 0.72 & \underline{\textbf{0.88}} \\
  & & Warning & 0.96 & 0.97 & 0.80 & 0.88 & 0.94 & 0.94 & \textbf{0.78} & 0.83 \\
\bottomrule
\end{tabular}
\caption{Comparison between execution rate of error recovery strategies.}
\label{tab:appendix_k4_feedback_exec}
\end{threeparttable}
\end{table}

\begin{table}[!ht]
\small
\setlength{\modelspacing}{2pt}
\setlength{\tabcolsep}{1.7pt} % Default value: 6pt
\setlength{\belowrulesep}{4pt}
\begin{threeparttable}
    \centering
    \begin{tabular}{cc l r rrr @{\quad} rrrr}
\toprule
\multirow{2}{*}{} & \multirow{2}{*}{} & \multirow{2}{*}{Feedback} & \multirow{2}{*}{O} & \multicolumn{3}{c}{Distraction} & \multicolumn{4}{c}{Counterfactual} \\
 & & & & E& L & T & $\text{O}_C$ & $\text{E}_C$& $\text{L}_C$ & $\text{T}_C$\\
\midrule
\multirow{6}{*}{\rotatebox{90}{Gemma-2}} & \multirow{3}{*}{\rotatebox{90}{9b}}
   & No recovery & \textbf{0.81} & \underline{\textbf{0.83}} & 0.70 & \underline{\textbf{0.91}} & \textbf{0.72} & \textbf{0.58} & \textbf{0.73} & 0.57 \\
 & & Error type & 0.77 & 0.78 & 0.69 & 0.76 & 0.64 & \textbf{0.58} & 0.69 & 0.59 \\
 & & Error message & 0.75 & 0.75 & \textbf{0.80} & 0.73 & 0.65 & 0.56 & 0.65 & \underline{\textbf{0.79}} \\
\cmidrule{2-11}
 & \multirow{3}{*}{\rotatebox{90}{27b}} 
   & No recovery & \underline{\textbf{0.85}} & \textbf{0.82} & 0.78 & \textbf{0.83} & \underline{\textbf{0.75}} & \underline{\textbf{0.71}} & \underline{\textbf{0.74}} & 0.54 \\
 & & Error type & 0.84 & 0.81 & 0.77 & 0.79 & 0.71 & 0.68 & 0.62 & 0.61 \\
 & & Error message & 0.83 & 0.79 & \textbf{0.79} & 0.76 & 0.71 & 0.70 & 0.72 & \textbf{0.64} \\
\midrule
\multirow{6}{*}{\rotatebox{90}{Mistral}} & \multirow{3}{*}{\rotatebox{90}{7B}} 
   & No recovery & 0.70 & \textbf{0.76} & 0.69 & 0.66 & 0.48 & 0.38 & \textbf{0.58} & 0.46 \\
 & & Error type & \textbf{0.73} & 0.64 & \textbf{0.75} & \textbf{0.69} & \textbf{0.54} & \textbf{0.49} & 0.55 & 0.51 \\
 & & Error message & 0.69 & 0.68 & 0.62 & \textbf{0.69} & 0.51 & 0.45 & 0.56 & \textbf{0.56} \\
\cmidrule{2-11}
 & \multirow{3}{*}{\rotatebox{90}{Small}} 
   & No recovery & \textbf{0.82} & \textbf{0.77} & \underline{\textbf{0.86}} & \textbf{0.81} & 0.58 & \textbf{0.63} & 0.52 & \textbf{0.69} \\
 & & Error type & 0.80 & 0.76 & 0.73 & \textbf{0.81} & \textbf{0.62} & \textbf{0.63} & \underline{\textbf{0.74}} & 0.67 \\
 & & Error message & 0.81 & \textbf{0.77} & 0.77 & 0.78 & 0.60 & 0.59 & 0.69 & 0.66 \\
\midrule
\multirow{8}{*}{\rotatebox{90}{Llama-3.1}} & \multirow{4}{*}{\rotatebox{90}{8B}} 
   & No recovery & \textbf{0.82} & \textbf{0.80} & 0.78 & 0.65 & 0.61 & 0.60 & 0.55 & \textbf{0.58} \\
 & & Error type & 0.81 & 0.75 & 0.76 & \textbf{0.71} & \textbf{0.67} & 0.54 & 0.55 & \textbf{0.58} \\
 & & Error message & 0.80 & 0.75 & \textbf{0.79} & 0.65 & 0.61 & 0.57 & \textbf{0.64} & 0.52 \\
  & & Warning & 0.79 & 0.75 & 0.70 & 0.69 & 0.54 & \textbf{0.61} & 0.51 & 0.39 \\
\cmidrule{2-11}
 & \multirow{4}{*}{\rotatebox{90}{70B}} 
   & No recovery & \textbf{0.82} & \textbf{0.80} & \textbf{0.80} & \textbf{0.83} & \textbf{0.66} & 0.63 & \textbf{0.64} & 0.61 \\
 & & Error type & 0.73 & 0.72 & 0.76 & 0.77 & 0.62 & 0.56 & 0.60 & 0.63 \\
 & & Error message & 0.72 & 0.71 & 0.75 & 0.75 & 0.64 & 0.60 & 0.54 & \textbf{0.66} \\
  & & Warning & 0.71 & 0.72 & 0.74 & 0.77 & 0.63 & \textbf{0.65} & 0.61 & \textbf{0.66} \\
\midrule
\multirow{4}{*}{\rotatebox{90}{GPT}} & \multirow{4}{*}{\rotatebox{90}{4o-mini}} 
   & No recovery & \underline{\textbf{0.85}} & \underline{\textbf{0.83}} & 0.80 & 0.82 & 0.64 & \textbf{0.64} & \textbf{0.63} & 0.57 \\
 & & Error type & 0.83 & 0.80 & 0.80 & 0.79 & 0.68 & 0.58 & 0.56 & 0.56 \\
 & & Error message & \underline{\textbf{0.85}} & 0.79 & \textbf{0.81} & \textbf{0.83} & 0.62 & 0.59 & 0.57 & 0.57 \\
  & & Warning & \underline{\textbf{0.85}} & 0.76 & 0.78 & 0.81 & \textbf{0.71} & 0.62 & 0.62 & \textbf{0.58} \\
\bottomrule
\end{tabular}
\caption{Comparison between  valid accuracy of error recovery strategies.}
\label{tab:appendix_k4_feedback_vacc}
\end{threeparttable}
\end{table} 

\clearpage
\section{Tautology corpus}
\label{app:tautology}
\begin{itemize}
    \item False is not true.
    \item True is not false.
    \item Not false is true.
    \item Not true is false.
    \item False and true is not true.
    \item False and not true is false.
    \item False and not false is false.
    \item Not true and true is false.
    \item Not true and false is false.
    \item True and false is not true.
    \item True and true is not false.
    \item True and not false is true.
    \item Not false and true is true.
    \item Not false and false is false.
    \item True or not true is true.
    \item True or true is true.
    \item False or not false is true.
    \item False or not true is false.
    \item Not true or false is not true.
    \item Not true or true is true.
    \item Not false or false is true.
    \item Not false or true is not false.
\end{itemize}

\section{Inference rules for counterfactual statements}
\textbf{Bidirectional dilemma} \\
$ \forall~ x ~(\mathsf{p}(x) \implies \mathsf{q}(x)) \land (\mathsf{r}(x) \implies \mathsf{s}(x)) \land (\mathsf{p}(a) \lor \neg \mathsf{s}(a)) \models (\mathsf{q}(a) \lor \neg \mathsf{r}(a))  $\\
Negated: $ \forall ~ x ~ (\mathsf{p}(x) \implies \neg\mathsf{q}(x)) \land (\mathsf{r}(x) \implies \mathsf{s}(x)) \land (\mathsf{p}(a) \lor \neg \mathsf{s}(a)) \models (\neg\mathsf{q}(a) \lor \neg \mathsf{r}(a))  $ \\
\textbf{Constructive dilemma} \\
$ \forall ~ x ~ ((\mathsf{p}(x) \implies \mathsf{q}(x)) \land (\mathsf{r}(x) \implies \mathsf{s}(x))) \land (\mathsf{p}(a) \lor \mathsf{r}(a)) \models (\mathsf{q}(a) \lor \mathsf{s}(a)) $ \\
Negated: $ \forall ~x ~ ((\mathsf{p}(x) \implies \neg\mathsf{q}(x)) \land (\mathsf{r}(x) \implies \mathsf{s}(x))) \land (\mathsf{p}(a) \lor \mathsf{r}(a)) \models (\neg\mathsf{q}(a) \lor \mathsf{s}(a)) $\\
\textbf{Disjunctive syllogism} \\
$ \forall ~ x ~ (\mathsf{p}(x) \lor \mathsf{q}(x)) \land \neg \mathsf{p}(a) \models \mathsf{q}(a)$ \\
Negated: $ \forall ~ x ~ (\mathsf{p}(x) \lor \neg\mathsf{q}(x)) \land \neg \mathsf{p}(a) \models \mathsf{q}(a)$\\
\textbf{Existential generalization} \\
$ \mathsf{p}(a) \models \exists ~ x~\mathsf{p}(x) $\\
Negated: $ \neg\mathsf{p}(a) \models \exists~ x~ \neg\mathsf{p}(x) $\\
\textbf{Hypothetical syllogism} \\
$ \forall ~ x ~ (\mathsf{p}(x) \implies \mathsf{q}(x)) \land (\mathsf{q}(x) \implies \mathsf{r}(x)) \models (\mathsf{p}(a) \implies \mathsf{r}(a)) $\\
Negated: $ \forall ~ x ~ (\neg\mathsf{p}(x) \implies \mathsf{q}(x)) \land (\mathsf{q}(x) \implies \mathsf{r}(x)) \models (\neg\mathsf{p}(a) \implies \mathsf{r}(a)) $\\
\textbf{Modus ponens} \\
$ \forall ~ x ~ (\mathsf{p}(x) \implies \mathsf{q}(x)) \land \mathsf{p}(a) \models \mathsf{q}(a)$ \\
Negated: $ \forall ~ x ~ (\mathsf{p}(x) \implies \neg\mathsf{q}(x)) \land \mathsf{p}(a) \models \neg\mathsf{q}(a)$ \\
\textbf{Modus tollens} \\
$ \forall ~ x ~(\mathsf{p}(x) \implies \mathsf{q}(x)) \land \neg \mathsf{q}(a) \models \neg \mathsf{p}(a)$ \\
Negated: $ \forall ~ x ~(\neg\mathsf{p}(x) \implies \mathsf{q}(x)) \land \neg \mathsf{q}(a) \models \mathsf{p}(a)$  \\
\textbf{Universal instantiation} \\
$ \forall ~ x ~ \mathsf{p}(x) \models \mathsf{p}(a) $\\
Negated: $ \forall ~ x ~ \neg\mathsf{p}(x) \models \neg\mathsf{p}(a) $ \\

\section{Prompts}
\label{app:prompts}
\subsection{Direct prompt}
\begin{lstlisting}
Given the context and question, answer the question and consider that not necessarily the whole context is relevant. Answer the question ONLY in 'yes' or 'no'. Please use the below format:
Context: [text with logical rules]
Question: [question based on context]
Answer: Yes/No
----
Context: If someone walks in the rain, they will get wet. Conversely, if someone exercises a lot, they will get fit. Leaders of a country for life are either a king or a queen. It is known that at least one of the following statements is true: (1) either John walks in the rain and (2) he will not get fit. It is possible that solely (1) is true, or solely (2) is true, or even both are true simultaneously.
Question: Can we say at least one of the following must always be true? (a) he will get wet and (b) he does not exercises a lot?
Reasoning steps:  1. John walks in the rain or he will not get fit 2. If John walks in the rain he will get wet. 3. If John will not get fit he cannot exercise a lot.
Answer: Yes
----
 Context: If a person leaves late, they will miss their train. In this particular situation, James left late.
Question: Does this entail that he will not miss his train?
Reasoning steps: 
 1. James left late 2. James left late, he will miss his train. 3. John misses his train contradicts that he will not miss his train.
Answer: No
----
 Context: It is known that one of the following options is true: someone goes to the office or someone goes home. However, Jill does not go to the office. If some pet in the office barks, then it is not dead.
Question: Does this imply that Jill goes home?
Reasoning steps:  1. Jill goes to the office or she goes home. 2. If Jill does not go to the office she must go home instead.
Answer: Yes
----
\end{lstlisting}
Regular expression for answer extraction: 
\begin{lstlisting}
r"(.*)(?P<answer>(yes)|(no))(.*)"
\end{lstlisting}

\subsection{CoT prompt}
\begin{lstlisting}
Task Description: Given the context and question, think step-by-step logically to answer the question and consider that not necessarily the whole context is relevant. Answer the question ONLY in 'yes' or 'no'. Please use the below format:
Context: [text with logical rules]
Question: [question based on context]
Reasoning steps: [generate step-by-step
reasoning]
Answer: Yes/No
----
Context: If someone walks in the rain, they will get wet. Conversely, if someone exercises a lot, they will get fit. Leaders of a country for life are either a king or a queen. It is known that at least one of the following statements is true: (1) either John walks in the rain and (2) he will not get fit. It is possible that solely (1) is true, or solely (2) is true, or even both are true simultaneously.
Question: Can we say at least one of the following must always be true? (a) he will get wet and (b) he does not exercises a lot?
Reasoning steps:  1. John walks in the rain or he will not get fit 2. If John walks in the rain he will get wet. 3. If John will not get fit he cannot exercise a lot.
Answer: Yes
----
 Context: If a person leaves late, they will miss their train. In this particular situation, James left late.
Question: Does this entail that he will not miss his train?
Reasoning steps: 
 1. James left late 2. James left late, he will miss his train. 3. John misses his train contradicts that he will not miss his train.
Answer: No
----
 Context: It is known that one of the following options is true: someone goes to the office or someone goes home. However, Jill does not go to the office. If some pet in the office barks, then it is not dead.
Question: Does this imply that Jill goes home?
Reasoning steps:  1. Jill goes to the office or she goes home. 2. If Jill does not go to the office she must go home instead.
Answer: Yes
----
\end{lstlisting}
Regular expression for answer extraction: 
\begin{lstlisting}
r"(.*)answer\:\s*(?P<answer>(yes)|(no))(.*)"
\end{lstlisting}

\subsection{Autoformalistion prompt}
\begin{lstlisting}[escapeinside={[}{]}]
Task Description: Given a problem description and a question. The task is to parse the problem and the question into first-order logic formulars. Follow exactly the given structure and consider that not necessarily the whole context is relevant.
The grammar of the first-order logic formular is defined as follows:
1) logical conjunction of expr1 and expr2: expr1 [$\land$] expr2
2) logical disjunction of expr1 and expr2: expr1 [$\lor$] expr2
3) logical exclusive disjunction of expr1 and expr2: expr1 [$\oplus$] expr2
4) logical negation of expr1: [$\neg$]expr1
5) expr1 implies expr2: expr1 [$\rightarrow$] expr2
6) expr1 if and only if expr2: expr1 [$\leftrightarrow$] expr2
7) logical universal quantification: [$\forall$]x
8) logical existential quantification: [$\exists$]x
----
Problem: If someone walks in the rain, they will get wet. Conversely, if someone exercises a lot, they will get fit. Leaders of a country for life are either a king or a queen. It is known that at least one of the following statements is true: (1) either John walks in the rain and (2) he will not get fit. It is possible that solely (1) is true, or solely (2) is true, or even both are true simultaneously.
Question: Can we say at least one of the following must always be true? (a) he will get wet and (b) he does not exercises a lot?
Predicates:
WalksInRain(x) ::: x walks in the rain.
GetWet(x) ::: x gets wet.
ExercisesALot(x) ::: x exercises a lot.
GetFit(x) ::: x gets fit.
Premises:
[$\forall$]x (WalksInRain(x) [$\rightarrow$] GetWet(x)) ::: If someone walks in the rain, they will get wet.
[$\forall$]x (ExercisesALot(x) [$\rightarrow$] GetFit(x)) ::: If someone exercises a lot, they will get fit.
(WalksInRain(john) [$\lor$] [$\neg$]GetFit(john)) ::: John walks in the rain or he will not get fit.
Conclusion:
GetWet(john) [$\wedge$] [$\neg$]ExercisesALot(john) ::: John will get wet or he does not exercises a lot.
----
Problem: If a person leaves late, they will miss their train. In this particular situation, James left late.
Question: Does this entail that he will not miss his train?
Predicates:
LeaveLate(x) ::: x leaves late.
MissTrain(x) ::: x misses their train.
Premises:
[$\forall$]x (LeaveLate(x) [$\rightarrow$] MissTrain(x)) ::: If a person leaves late, they will miss their train.
LeaveLate(james) ::: James leavs late.
Conclusion:
[$\neg$]MissTrain(james) ::: James does not miss his train.
----
Problem: It is known that one of the following options is true: someone goes to the office or someone goes home. However, Jill does not go to the office. If some pet in the office barks, then it is not dead.
Question: Does this imply that Jill goes home?
Predicates:
GoesToOffice(x) ::: x goes to the office.
GoesHome(x) ::: x goes home.
Premises:
[$\forall$]x (GoesToOffice(x) [$\lor$] GoesHome(x)) ::: Either someone goes to the office or someone goes home.
[$\neg$]GoesToOffice(jill) ::: Jill does not go to the office.
Conclusion:
GoesHome(jill) ::: Jill goes home.
----
\end{lstlisting}